\documentclass[11pt]{article}

\PassOptionsToPackage{hyperfootnotes=false}{hyperref}
\usepackage[preprint]{acl}

\usepackage{times}
\usepackage{latexsym}

\usepackage[T1]{fontenc}
\usepackage[utf8]{inputenc}
\usepackage{microtype}
\usepackage{inconsolata}
\usepackage{url}
\usepackage{booktabs}
\usepackage{graphicx}
\usepackage{amsfonts}
\usepackage{amsmath}
\usepackage{nicefrac}
\usepackage{xcolor}
\usepackage{xspace}
\usepackage{multirow}
\usepackage{multicol}
\usepackage{array}
\usepackage{algorithm}
\usepackage{algpseudocode}

\newcommand{\ours}{PLC-DPO\xspace}
\newcommand{\method}{Posterior Label Correction DPO\xspace}
\newcommand{\pclean}{q_{\mathrm{clean}}}
\newcommand{\pflip}{q_{\mathrm{flip}}}
\newcommand{\ptie}{q_{\mathrm{tie}}}
\newcommand{\Lclean}{\mathcal{L}_{\mathrm{clean}}}
\newcommand{\Lflip}{\mathcal{L}_{\mathrm{flip}}}
\newcommand{\Ltie}{\mathcal{L}_{\mathrm{tie}}}
\newcommand{\Lplc}{\mathcal{L}_{\mathrm{PLC}}}
\newcommand{\Ldpo}{\mathcal{L}_{\mathrm{DPO}}}
\newcommand{\mseq}{m_{\mathrm{seq}}}
\newcommand{\stopgrad}{\operatorname{stopgrad}}
\newcommand{\softplus}{\operatorname{softplus}}
\newcommand{\algcomment}[1]{\hfill{\footnotesize\textcolor{blue}{$\triangleright$ #1}}}

\title{\ours{}: Posterior Label Correction in\\Noisy and Ambiguous Preference Optimization}
\newcommand{\correspondingmark}{\textsuperscript{\textdagger}}

\author{Boryeong Cho \\
  KAIST AI \\
  \texttt{venntum@kaist.ac.kr} \\\And
  Sumyeong Ahn\correspondingmark \\
  KENTECH \\
  \texttt{sumyeongahn@kentech.ac.kr} \\\And
  Se-Young Yun\correspondingmark \\
  KAIST AI \\
  \texttt{yunseyoung@kaist.ac.kr}}

\begin{document}
\maketitle
\makeatletter
\ifacl@anonymize\else
\begingroup
\renewcommand{\thefootnote}{\fnsymbol{footnote}}
\footnotetext[2]{Corresponding authors.}
\endgroup
\footnotetext[1]{\url{https://github.com/VennTum99/PLC-DPO}}
\fi
\makeatother

\begin{abstract}
Direct Preference Optimization (DPO) simplifies alignment through pairwise comparisons but assumes all observed preferences are reliable. Real data often violates this assumption, leading to reversed, weak, or ambiguous labels that cause harmful policy updates. To address this, we propose \method{} (\ours{}) to robustly optimize preferences by routing each pair's training signal as a clean, flip, or tie case. The key idea is to use the calibrated policy-reference margin as online evidence to take appropriate correction actions. This reframes noisy preference learning as actively correcting supervision direction and strength rather than merely filtering suspicious examples. Across 57 dataset--model--benchmark cells, \ours{} obtains the best mean win rate against DPO (60.5 vs. 55.5 for the next-best method). Injected-noise and tie stress tests, human disagreement analysis, and self-confirmation diagnostics further show that the routing remains stable and distinguishes flipped from weakly directional pairs.
\end{abstract}

\section{Introduction}
\label{sec:introduction}

\begin{figure*}[t]
\centering
\includegraphics[width=0.8\textwidth]{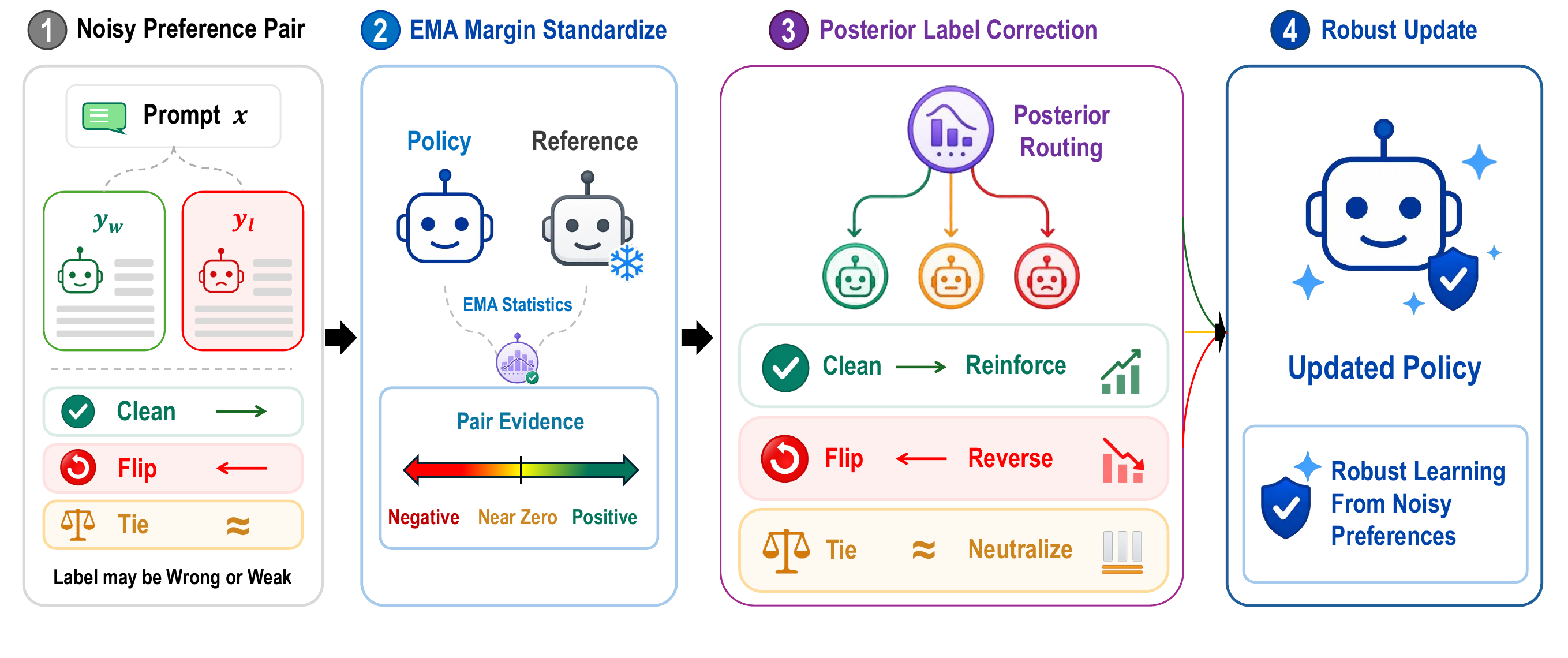}
\captionsetup{aboveskip=2pt}
\caption{Overview of \ours. Preference pairs may be clean, flipped, or weakly directional. \ours{} uses the standardized policy-reference margin as evidence for posterior routing, then reinforces, reverses, or neutralizes each pair to enable robust learning from noisy preferences.}
\label{fig:method-overview}
\end{figure*}

Direct Preference Optimization (DPO)~\citep{rafailov2023dpo} is now a standard objective for the offline alignment of large language models. While traditional reinforcement learning from human feedback (RLHF) requires training a separate reward model~\citep{ouyang2022rlhf,schulman2017ppo}, DPO eliminates this step. It fits a Bradley-Terry preference model directly using the policy-reference log-ratio~\citep{bradleyterry1952model}. This closed-form pairwise loss is simple, scalable, and easily applied to large preference datasets.

This simplicity is also its vulnerability. DPO assumes every observed preference $y_w \succ y_l$ is a completely reliable target. In reality, preference labels suffer from annotator disagreement, model-judge biases, and superficial cues like length or confident wording~\citep{zhang2025diverging,zheng2023llmasajudge,wang2024notfair}. Furthermore, true preferences are not always strictly directional. Two responses might be similarly good, equally flawed, or too marginally different to provide a stable binary signal. By treating all labels as absolute ground truth, DPO actively reinforces incorrect optimization directions when faced with noisy or ambiguous data.

Recent studies address this uncertainty indirectly. Robust-loss variants typically assume a global noise rate and apply uniform corrections across all pairs~\citep{chowdhury2024robustdpo,wu2025drdpo}. Data filtering and curriculum methods discard suspicious data entirely, which wastes useful signals that could be extracted by recalibrating flipped labels~\citep{gao2025selectivedpo,liang2025ropo}. Alternatively, latent-quality approaches estimate absolute response scores to infer pairwise preferences. Although effective, these pointwise methods operate indirectly, unlike the core objective of directly optimizing relative pairwise directions.

We propose \method{} (\ours{})\footnotemark[1], an online robust preference optimization objective that directly models the observed pair label as a latent clean, flip, or tie state. A clean state means the observed direction should be reinforced. A flip state means the direction should be reversed. A tie state means the pair should not induce a strong directional gradient. \ours{} estimates a posterior-like routing distribution over these states from the policy-reference preference margin, calibrates this signal with an exponential moving average, stops gradients through the routing weights, and uses them to mix forward DPO, reversed DPO, and tie-regularizing losses. A warm-up schedule and routing-confidence gate keep the method close to DPO until the correction signal becomes informative.

Our contributions are as follows:
\begin{itemize}
    \item We introduce \ours, an online robust preference optimization objective that models each observed pairwise preference label as a latent clean, flip, or tie state and estimates a posterior-like routing distribution from the calibrated policy-reference preference margin.
    \item We derive a routing-mixed objective that softly combines forward DPO, reversed DPO, and tie-regularizing losses. Unlike response-level latent-quality routing or external reward-model filtering, \ours{} directly corrects the pair label consumed by DPO and requires no additional supervision.
    \item We propose a stable training recipe based on EMA calibration, routing, warm-up, and confidence-gated mixing, enabling label correction without early-training instability.
    \item We evaluate \ours{} across models, preference datasets, alignment benchmarks, label-noise and tie stress tests, an independent human-disagreement evaluation set, routing diagnostics, and ablations. Across 57 dataset--model--benchmark cells, \ours{} achieves the highest mean win rate and the second-highest worst-cell result among the robust baselines.
\end{itemize}

\section{Related Work}
\label{sec:related-work}

\begin{algorithm*}[t]
\centering
\begin{minipage}{\textwidth}
\caption{\ours{} Training Step}
\label{alg:plc-dpo}
\small
\begin{algorithmic}[1]
    \State \textbf{Input} Mini-Batch $\mathcal{B}=\{(x_i,y_{w,i},y_{l,i})\}_{i=1}^{B}$, policy $\pi_\theta$, reference $\pi_{\mathrm{ref}}$, EMA state $(\mu,v)$, step $t$, total steps $T$
    \State \textbf{Hyperparameters} direction temperature $\tau_{\mathrm{dir}}$, tie temperature $\tau_{\mathrm{tie}}$, maximum correction strength $\gamma_{\max}$, confidence power $\kappa$
    \State \textbf{Fixed settings} DPO scale $\beta$, EMA decay $\alpha$, minimum std $\sigma_{\min}$, warm-up schedule, initial state prior $\pi^0$
    \Statex
    \State Compute policy and reference log-probabilities for $(y_{w,i},y_{l,i})$ \algcomment{forward pass}
    \State ${\mseq}_i \gets \beta\!\left[\log\frac{\pi_\theta(y_{w,i}\mid x_i)}{\pi_{\mathrm{ref}}(y_{w,i}\mid x_i)}-\log\frac{\pi_\theta(y_{l,i}\mid x_i)}{\pi_{\mathrm{ref}}(y_{l,i}\mid x_i)}\right]$, \quad ${\Ldpo}_i \gets -\log\sigma({\mseq}_i)$ \algcomment{DPO backbone}
    \State $\tilde m_i \gets \stopgrad({\mseq}_i)$ \algcomment{routing margin signal}
    \State $\bar m \gets \frac{1}{B}\sum_i \tilde m_i$, \quad $s_m^2 \gets \frac{1}{B}\sum_i(\tilde m_i-\bar m)^2$ \algcomment{batch statistics}
    \State $\mu \gets \alpha\mu+(1-\alpha)\bar m$, \quad $v \gets \alpha v+(1-\alpha)s_m^2$, \quad $z_i \gets \frac{\tilde m_i-\mu}{\max(\sqrt{v},\sigma_{\min})}$ \algcomment{online calibration}
    \State $\ell_{i,\mathrm{clean}}\gets\log \pi^0_{\mathrm{clean}}+z_i/\tau_{\mathrm{dir}}$, \quad $\ell_{i,\mathrm{flip}}\gets\log \pi^0_{\mathrm{flip}}-z_i/\tau_{\mathrm{dir}}$, \quad $\ell_{i,\mathrm{tie}}\gets\log \pi^0_{\mathrm{tie}}-|z_i|/\tau_{\mathrm{tie}}$ \algcomment{data-dependent evidence}
    \State $({\pclean}_i,{\pflip}_i,{\ptie}_i) \gets \stopgrad\!\left(\operatorname{softmax}(\ell_{i,\mathrm{clean}},\ell_{i,\mathrm{flip}},\ell_{i,\mathrm{tie}})\right)$ \algcomment{routing weights}
    \State ${\Lclean}_i \gets -\log\sigma({\mseq}_i)$, \quad ${\Lflip}_i \gets -\log\sigma(-{\mseq}_i)$, \quad ${\Ltie}_i \gets \softplus(|{\mseq}_i|)$ \algcomment{state losses}
    \State ${\Lplc}_i \gets {\pclean}_i{\Lclean}_i+{\pflip}_i{\Lflip}_i+{\ptie}_i{\Ltie}_i$ \algcomment{routing-corrected loss}
    \State $C_i \gets g_\kappa({\pclean}_i,{\pflip}_i,{\ptie}_i)$ \algcomment{pair-level routing confidence}
    \If{$t < \rho_{\mathrm{warm}}T$}
        \State $\gamma_t \gets 0$ \algcomment{warm-up}
    \Else
        \State $\gamma_t \gets \min(\gamma_{\max},\mathrm{Schedule}(t))$ \algcomment{increase correction strength}
    \EndIf
    \State $w_i \gets \gamma_t C_i$ \algcomment{actual correction weight}
    \State $\mathcal{L}_i \gets (1-w_i){\Ldpo}_i+w_i{\Lplc}_i$, \quad $\theta \gets \operatorname{OptimizerStep}\!\left(\theta,\nabla_\theta\frac{1}{B}\sum_i\mathcal{L}_i\right)$ \algcomment{policy update}
\end{algorithmic}
\end{minipage}
\end{algorithm*}

\paragraph{Preference optimization.}
RLHF aligns a language model by training an explicit reward model on pairwise preferences and then optimizing the policy with PPO~\citep{ouyang2022rlhf,schulman2017ppo}. DPO reparameterizes the optimal policy through the policy-reference log-ratio and yields a closed-form pairwise loss~\citep{rafailov2023dpo}. Subsequent methods modify the preference objective, supervision format, or reward parameterization. KTO optimizes from desirable and undesirable generations without requiring paired comparisons~\citep{ethayarajh2024kto}. SimPO uses a reference-free and length-normalized preference reward~\citep{meng2024simpo}. RSO improves preference optimization through rejection sampling~\citep{liu2024rso}. These objective-level advances improve how preferences are optimized, but they typically keep the observed pair direction fixed once a preference pair is constructed. As a result, they do not directly address cases where the pair label itself should be reversed or treated as non-directional.

\paragraph{Robust preference optimization under noisy labels.}
Recent works explore preference optimization under noisy, weak, or biased labels. Label-smoothed and robust DPO losses reduce overconfidence or debias targets against random flips~\citep{mitchell2023cdpo,chowdhury2024robustdpo}. Dr.DPO uses a distributionally robust framework to control pairwise reliability~\citep{wu2025drdpo}. ROPO combines a noise-aware loss with iterative filtering~\citep{liang2025ropo}, while $\gamma$-PO adopts pair-specific dynamic margins~\citep{sun2025gammapo}. RE-PO applies an EM-style posterior to reweight observed and reversed directions across preference losses~\citep{cao2026repo}. Semi-supervised variants similarly estimate trustworthiness to downweight or smooth uncertain updates~\citep{liu2026semidpo}. Most of these methods address noise through reliability weighting, smoothing, filtering, or dynamic margins. They leave less explicit the case where a pair is preference-uninformative rather than merely clean or flipped, and therefore should avoid inducing a strong directional gradient.
\section{Posterior Label Correction DPO}
\label{sec:method}

\ours{} treats noisy preference optimization as an online latent-label problem. The latent label is not the quality of either response in isolation, state of the pairwise direction that DPO consumes. This section defines the standard DPO margin, derives the calibrated clean/flip/tie routing distribution, and gives the final confidence-gated objective.

\subsection{Problem Setup}
\label{sec:problem-setup}

Let $x$ be a prompt and let $(y_w,y_l)$ denote the observed chosen and rejected responses. A trainable policy $\pi_\theta$ is initialized from a supervised fine-tuned model, and $\pi_{\mathrm{ref}}$ is a frozen reference policy. Standard DPO optimizes a Bradley-Terry model with the sequence-level margin
\begin{equation}
\begin{aligned}
\mseq(x,y_w,y_l)
&= \beta
\bigg[
\log \frac{\pi_\theta(y_w \mid x)}
{\pi_{\mathrm{ref}}(y_w \mid x)}
\\[-0.2em]
&\qquad
-
\log \frac{\pi_\theta(y_l \mid x)}
{\pi_{\mathrm{ref}}(y_l \mid x)}
\bigg],
\end{aligned}
\label{eq:mseq}
\end{equation}
where $\beta$ controls the reward scale. The standard DPO loss~\citep{rafailov2023dpo} is
\begin{equation}
\Ldpo(\theta) = -\log \sigma(\mseq).
\label{eq:dpo}
\end{equation}
DPO is effective when the observed direction is reliable. When the pair is flipped or non-directional, it turns that error into a direct gradient on the policy.

\subsection{Pair Label as a Latent State}
\label{sec:latent-state}

We introduce a latent state $s \in \{\mathrm{clean},\mathrm{flip},\mathrm{tie}\}$ for every observed pair. The clean state means the observed ordering $y_w \succ y_l$ should be reinforced. The flip state means the opposite ordering should be learned. The tie state means the pair should not induce a strong directional preference gradient. In this paper, tie denotes a pair where the observed direction is not sufficiently informative, either because both responses are similarly good, both are similarly bad, or the current policy-reference margin provides insufficient directional evidence.

This choice targets the variable used by DPO. Response-level latent-quality methods ask whether $y_w$ and $y_l$ are individually good or bad, then infer a pair action from those two estimates. \ours{} instead asks whether the observed pair direction is clean, flipped, or non-directional, then maps the answer directly to a loss.

\subsection{Margin-Based Routing Distribution}
\label{sec:posterior}

The routing distribution is estimated from the same pairwise margin that DPO uses for optimization, but the margin is first detached and calibrated online. For each example $i$, let
\begin{equation}
\tilde m_i = \stopgrad({\mseq}_i).
\end{equation}
The stop-gradient operation makes the routing weights an assignment signal for the current update rather than an additional path through which the policy can reduce the loss by changing its own label assignment.

Because the scale of $\tilde m_i$ changes during training, \ours{} maintains an exponential moving average (EMA) of the batch mean and variance over the training. For a batch at step $t$, let $\bar m_t$ and $s_t^2$ denote the mean and variance of $\{\tilde m_i\}_{i=1}^{B}$. We update
\begin{align}
\mu_t &= \alpha \mu_{t-1} + (1-\alpha)\bar m_t, \\
v_t &= \alpha v_{t-1} + (1-\alpha)s_t^2.
\end{align}
The calibrated margin for pair $i$ is
\begin{equation}
z_i = \frac{\tilde m_i-\mu_t}{\max(\sqrt{v_t},\sigma_{\min})}.
\label{eq:standardized-margin}
\end{equation}
Large positive $z_i$ means that the current policy-reference signal agrees with the observed label, large negative $z_i$ means it contradicts the label, and values near zero provide weak directional evidence.

We convert the calibrated margin into three energy scores.
\begin{align}
\ell_{\mathrm{clean}} &= \log \pi^0_{\mathrm{clean}} + z/\tau_{\mathrm{dir}}, \\
\ell_{\mathrm{flip}} &= \log \pi^0_{\mathrm{flip}} - z/\tau_{\mathrm{dir}}, \\
\ell_{\mathrm{tie}} &= \log \pi^0_{\mathrm{tie}} - |z|/\tau_{\mathrm{tie}}.
\end{align}
Here $\tau_{\mathrm{dir}}$ controls how sharply signed evidence separates clean from flip, while $\tau_{\mathrm{tie}}$ controls how quickly tie evidence decays away from zero. The constants $\pi^0_s$ act as initial state preferences. They are not intended to define a normalized generative model for $z$. Instead, they provide an energy-based, posterior-like routing score for choosing the training action supported by the current calibrated margin. We then normalize these scores with a softmax.
\begin{equation}
q_s
= \frac{\exp(\ell_s)}
{\sum_{s'} \exp(\ell_{s'})}.
\label{eq:posterior}
\end{equation}
The resulting $(\pclean,\pflip,\ptie)$ are therefore interpreted as differentiable routing weights over clean, flip, and tie actions, not as calibrated probabilities from a fully specified data-generating model.

\subsection{State-Conditional Losses}
\label{sec:state-losses}

\begin{table*}[t]
  \centering
  \small
  \caption{Main win-rate results against the one-epoch DPO baseline trained on UltraFeedback Binarized. The SFT row shows the model initialization before preference optimization. \ours{} uses the default recipe, and all methods share the same one-epoch training budget. Abbreviations denote their respective evaluation sets, with UFB, Alpaca, and Alpaca2 corresponding to UltraFeedback, AlpacaEval, and AlpacaEval 2. \textbf{Bold} and \underline{underline} denote the best and second-best results, respectively.}
  \label{tab:main-results}
  \resizebox{\textwidth}{!}{%
  \begin{tabular}{clccccccc}
    \toprule
    Model & Method & UFB $\uparrow$ & Alpaca $\uparrow$ & Alpaca2 $\uparrow$ & MT-Bench $\uparrow$ & Vicuna $\uparrow$ & Evol-Instruct $\uparrow$ & HH-RLHF $\uparrow$ \\
    \midrule
    \multirow{10}{*}{\begin{tabular}[c]{@{}c@{}}\textbf{Qwen2.5}\\\textbf{1.5B}\end{tabular}} & SFT & 21.60 & 14.41 & 13.79 & 21.25 & 5.00 & 15.65 & 32.67 \\
    & cDPO & 46.83 & 43.11 & 44.78 & \textbf{55.00} & 48.75 & 46.40 & 39.07 \\
    & rDPO & 50.42 & 48.32 & 49.69 & 52.50 & 54.37 & 51.55 & 47.64 \\
    & KTO-Pair & 47.52 & 50.50 & 48.57 & 48.12 & 48.75 & 53.29 & 45.90 \\
    & RSO & 50.88 & 50.43 & 50.31 & 50.00 & 51.25 & 54.72 & \underline{54.29} \\
    & $\gamma$-PO & 48.88 & 49.63 & 51.37 & \underline{53.12} & 52.50 & 49.88 & 52.80 \\
    & Dr.DPO & \underline{51.20} & 48.70 & 47.08 & 49.38 & 53.12 & 51.93 & 53.85 \\
    & ROPO & 49.45 & \underline{53.29} & \underline{52.55} & 45.00 & 58.75 & \underline{56.02} & 44.04 \\
    & RE-PO & 49.55 & 52.61 & 51.61 & 50.62 & \underline{61.25} & 53.23 & \textbf{54.84} \\
    & PLC-DPO & \textbf{52.48} & \textbf{55.03} & \textbf{57.70} & 42.50 & \textbf{66.88} & \textbf{58.82} & 45.78 \\

    \midrule

    \multirow{10}{*}{\begin{tabular}[c]{@{}c@{}}\textbf{Phi2}\\\textbf{2.7B}\end{tabular}} & SFT & 34.75 & 25.90 & 26.15 & 30.63 & 12.50 & 30.93 & 45.34 \\
    & cDPO & 49.18 & 48.63 & 45.40 & 38.75 & 43.12 & 48.57 & 51.49 \\
    & rDPO & 47.52 & 46.89 & 45.28 & 47.50 & 44.38 & 48.63 & 50.37 \\
    & KTO-Pair & \underline{56.47} & 58.63 & 57.64 & \underline{59.38} & 62.50 & 56.89 & 45.47 \\
    & RSO & 55.83 & 58.51 & 59.75 & 47.50 & 65.62 & \underline{58.01} & \textbf{56.15} \\
    & $\gamma$-PO & 47.93 & 52.30 & 52.36 & 48.75 & 51.88 & 49.19 & 52.98 \\
    & Dr.DPO & 47.67 & 49.69 & 47.52 & 51.88 & 55.62 & 51.18 & \underline{54.10} \\
    & ROPO & 55.95 & \underline{59.44} & \underline{60.75} & \textbf{61.25} & \underline{73.75} & \textbf{59.13} & 49.75 \\
    & RE-PO & 49.90 & 49.13 & 48.82 & 55.00 & 53.12 & 50.93 & 51.93 \\
    & PLC-DPO & \textbf{56.83} & \textbf{60.68} & \textbf{61.24} & 53.12 & \textbf{77.50} & 56.58 & 50.06 \\

    \midrule

    \multirow{10}{*}{\begin{tabular}[c]{@{}c@{}}\textbf{Qwen2.5}\\\textbf{7B}\end{tabular}} & SFT & 16.18 & 8.39 & 9.57 & 19.38 & 4.38 & 9.75 & 11.68 \\
    & cDPO & 48.62 & 41.30 & 46.34 & 54.37 & 45.00 & 42.24 & 40.81 \\
    & rDPO & 47.27 & 44.53 & 48.82 & 53.12 & 55.00 & 46.21 & 43.35 \\
    & KTO-Pair & 49.55 & 49.69 & 52.80 & 52.50 & 57.50 & 49.25 & 43.73 \\
    & RSO & 55.00 & 54.91 & 56.46 & 56.88 & \underline{75.00} & 55.34 & \underline{53.04} \\
    & $\gamma$-PO & 50.88 & 48.45 & 51.43 & 52.50 & 51.25 & 50.50 & 46.02 \\
    & Dr.DPO & 48.68 & 47.45 & 47.27 & 50.62 & 53.12 & 47.14 & 37.58 \\
    & ROPO & \underline{58.60} & \textbf{58.70} & \underline{58.94} & \underline{58.75} & 71.88 & \underline{59.94} & 48.32 \\
    & RE-PO & 43.77 & 41.49 & 42.55 & 51.25 & 48.12 & 44.16 & 36.52 \\
    & PLC-DPO & \textbf{58.80} & \underline{58.14} & \textbf{61.37} & \textbf{59.38} & \textbf{76.88} & \textbf{62.98} & \textbf{65.40} \\

    \bottomrule
  \end{tabular}
  }
\end{table*}

Given a loss margin $m$, the three state-conditional losses are
\begin{align}
\Lclean &= -\log \sigma(m), \label{eq:lclean}\\
\Lflip &= -\log \sigma(-m), \label{eq:lflip}\\
\Ltie &= \softplus\left(|m|\right). \label{eq:ltie}
\end{align}
In our main objective, $m=\mseq$, so the clean and flip losses operate on the same sequence-level margin as standard DPO. $\Lclean$ reinforces the observed direction, $\Lflip$ reverses it, and $\Ltie$ discourages large directional margins for pairs assigned to the tie state.

We stop gradients through the routing distribution.
\begin{equation}
\bar q_s = \stopgrad(q_s).
\label{eq:stopgrad}
\end{equation}
This prevents the policy from changing the state assignment and the state-conditional objective in the same update.

The routing-corrected loss is
\begin{equation}
\Lplc
= \bar q_{\mathrm{clean}}\Lclean
+ \bar q_{\mathrm{flip}}\Lflip
+ \bar q_{\mathrm{tie}}\Ltie.
\label{eq:lplc}
\end{equation}

\subsection{Warm-Up and Confidence-Gated Mixing}
\label{sec:gating}

Routing distributions are initially unreliable at the beginning of training due to weak policy-reference margins and uncalibrated EMA. Therefore, \ours{} trains with standard DPO during warm-up fraction $\rho_{\mathrm{warm}}$ of the total. Afterward, a schedule $\gamma_t$ increases the correction strength to $\gamma_{\max}$.

We also gate each pair using detached routing weights to estimate routing confidence. We define a confidence functional $g_\kappa: \Delta^2 \to [0,1]$ that is small for near-uniform distributions and large when a dominant latent state emerges. In our experiments, we use normalized maximum routing confidence
\begin{equation}
C(\bar q) =
g_\kappa(\bar q)
=
\left(
\frac{\max_s \bar q_s - 1/3}{2/3}
\right)^\kappa,
\label{eq:confidence}
\end{equation}
where $\kappa$ controls the deferral of low-confidence assignments. This correctly assigns zero confidence to a uniform distribution and unit confidence to a degenerate one. While entropy-based confidence is a natural alternative, it is overly conservative by penalizing the full distributional spread. Our max-weight gate instead ties confidence directly to the most likely correction action. The final loss is
\begin{equation}
\mathcal{L}(\theta)
=
(1-\gamma_t C(\bar q))\Ldpo
+ \gamma_t C(\bar q)\Lplc.
\label{eq:total}
\end{equation}

\subsection{Training Step}
\label{sec:algorithm}

\paragraph{Algorithmic summary.}
Algorithm~\ref{alg:plc-dpo} summarizes one mini-batch update. The algorithm follows the derivation above by computing the DPO margin, detaching and calibrating it to estimate the clean/flip/tie routing distribution, forming the state-conditional losses, and blending the routing-corrected objective with standard DPO through warm-up and confidence gating.

\paragraph{Implementation choices.}
The method introduces four objective-level controls. $\tau_{\mathrm{dir}}$ and $\tau_{\mathrm{tie}}$ determine how margin evidence is converted into routing mass, while $\gamma_{\max}$ and $\kappa$ determine how strongly confident corrections enter the final loss. The EMA decay $\alpha$ controls how quickly the streaming margin mean and variance follow the current training distribution. We keep calibration and scheduling settings fixed within each reported recipe, with values provided in Appendix~\ref{app:experimental-details} and component-level effects evaluated in Section~\ref{sec:ablations}.

\section{Experiments}
\label{sec:experiments}

We evaluate \ours{} around six questions. Q1 asks whether direct pair-label correction improves standard alignment quality over DPO and competitive preference-optimization baselines. Q2 asks whether these gains transfer across base models and preference datasets. Q3 asks whether the method remains robust when preference labels are flipped. Q4 asks whether the inferred routing states respond to controlled data pathologies as noise changes. Q5 asks whether the tie state responds to synthetic and human-annotated ambiguity. Q6 asks which components of the correction objective are necessary.

\subsection{Experimental Setup}
\label{sec:experimental-setup}

\paragraph{Models.}
We start from SFT models trained on \texttt{UltraChat-200k}~\citep{ding2023ultrachat}: \texttt{Qwen2.5-1.5B}, \texttt{Qwen2.5-7B}~\citep{qwen2025qwen25}, and \texttt{Phi-2-2.7B}~\citep{javaheripi2023phi2}. The generalization study additionally uses \texttt{Llama-3-8B}~\citep{grattafiori2024llama3} and \texttt{Mistral-7B}~\citep{jiang2023mistral7b}. Full training details are provided in Appendix~\ref{app:experimental-details}.

\paragraph{Baselines.}
We compare with SFT, standard DPO~\citep{rafailov2023dpo}, cDPO or label-smoothed DPO~\citep{mitchell2023cdpo}, rDPO~\citep{park2024rdpo}, KTO-Pair~\citep{ethayarajh2024kto}, RSO~\citep{liu2024rso}, and recent noisy-preference baselines including Dr.DPO~\citep{wu2025drdpo}, ROPO~\citep{liang2025ropo}, $\gamma$-PO~\citep{sun2025gammapo}, and RE-PO~\citep{cao2026repo}. Implementation details are in Appendix~\ref{app:experimental-details}.

\paragraph{Evaluation.}
We report pairwise win rates against the corresponding DPO baseline across UltraFeedback~\citep{cui2024ultrafeedback}, AlpacaEval, AlpacaEval 2~\citep{li2023alpacaeval}, MT-Bench~\citep{zheng2023llmasajudge}, Vicuna~\citep{chiang2023vicuna}, Evol-Instruct~\citep{xu2024evolinstrcut}, and HH-RLHF~\citep{bai2022hhrlhf} evaluation sets. Unless otherwise specified, \texttt{Skywork-Reward-V2-Llama-3.1-8B}~\citep{liu2026skyworkrewardv2} serves as the judge model. These experiments were conducted using single-run greedy decoding. For the main experiments, \ours{} uses the aggressive preset as its default recipe. Appendix~\ref{app:plc-recipe-sensitivity} compares this choice with other presets.

\begin{table}[t]
  \centering
  \small
  \caption{Commercial-model judge validation for Qwen2.5-7B. Claude Sonnet 4.6 scores each response independently and also judges pairwise comparisons against the corresponding DPO baseline.}
  \label{tab:commercial-judge-validation}
  \resizebox{0.8\columnwidth}{!}{%
  \begin{tabular}{lcccc}
    \toprule
    \multirow{2}{*}{Method} & \multicolumn{2}{c}{AlpacaEval 2} & \multicolumn{2}{c}{Vicuna} \\
    \cmidrule(lr){2-3}\cmidrule(lr){4-5}
    & Single & WR & Single & WR \\
    \midrule
    DPO & 5.585 & 50.00 & 6.600 & 50.00 \\
    rDPO & 5.620 & 51.00 & 6.588 & 48.75 \\
    $\gamma$-PO & 5.635 & 51.25 & 6.713 & 52.50 \\
    ROPO & \underline{5.785} & \underline{56.50} & \underline{6.725} & \underline{55.00} \\
    PLC-DPO & \textbf{5.795} & \textbf{56.75} & \textbf{6.900} & \textbf{61.25} \\
    \bottomrule
  \end{tabular}
  }
  \vspace{-10pt}
\end{table}

\subsection{Main Alignment Results}
\label{sec:main-results}

Table~\ref{tab:main-results} reports the main alignment results using a fixed default \ours{} recipe across all models and evaluation sets. \ours{} achieves the highest performance on most metrics for Qwen2.5-7B, showing particularly large gains on AlpacaEval 2, Vicuna, Evol-Instruct, and HH-RLHF. For Qwen2.5-1.5B and Phi-2-2.7B, the gains remain strong on AlpacaEval, AlpacaEval 2, and Vicuna, although ROPO and RE-PO are competitive on other splits. Notably, \ours{} yields greater improvements on larger models. This suggests that stronger base models inherently provide more accurate margin measurements for reliable routing. Finally, Appendix~\ref{app:plc-recipe-sensitivity} presents ablations across different recipe configurations.

\subsection{Generalization Across Models and Datasets}
\label{sec:generalization}

\begin{table*}[t]
  \centering
  \small
  \caption{Generalization with a single fixed recipe. (a) Mean win rate across 7 benchmarks against same-data DPO after training on clean UltraFeedback across different base models. (b) Mean and worst win rate over all 57 dataset--model--benchmark cells from UltraFeedback, HH-Golden, Nectar-60k, and ORPO-mix-40k. Full cell-level results and per-dataset averages are reported in Appendix~\ref{app:full-57-cell-results}.}
  \label{tab:extended-generalization}
  \begin{minipage}[t]{0.70\textwidth}
    \centering
    \textbf{(a) Clean UltraFeedback across different base models}\par\vspace{0.55em}
    \resizebox{0.90\linewidth}{!}{%
    \begin{tabular}{lcccccc}
      \toprule
      Model & rDPO & RE-PO & Dr.DPO & $\gamma$-PO & ROPO & \ours{} \\
      \midrule
      Qwen2.5-7B & 48.3 & 44.0 & 47.4 & 50.1 & \underline{59.3} & \textbf{63.3} \\
      Llama-3-8B & 43.8 & 51.2 & 54.0 & 49.7 & \textbf{60.4} & \underline{59.8} \\
      Mistral-7B & 44.1 & 48.9 & 49.3 & 49.1 & \underline{59.1} & \textbf{59.2} \\
      \bottomrule
    \end{tabular}}
  \end{minipage}\hfill
  \begin{minipage}[t]{0.23\textwidth}
    \centering
    \textbf{(b) Aggregate over 57 cells}\par\vspace{0.55em}
    \resizebox{\linewidth}{!}{%
    \begin{tabular}{lcc}
      \toprule
      Method & Mean $\uparrow$ & Worst $\uparrow$ \\
      \midrule
      rDPO & \underline{55.5} & 36.9 \\
      RE-PO & 49.2 & 36.5 \\
      Dr.DPO & 50.3 & 37.6 \\
      $\gamma$-PO & 49.7 & \textbf{42.5} \\
      ROPO & 46.7 & 12.7 \\
      \ours{} & \textbf{60.5} & \underline{41.2} \\
      \bottomrule
    \end{tabular}}
  \end{minipage}
\end{table*}

We train all six robust objectives and standard DPO from scratch using a single fixed recipe across every dataset--model pair, evaluating each method against the baseline DPO model. On clean UltraFeedback, \ours{} achieves a mean win rate of 60.7 across the 21 model--benchmark cells, followed by ROPO at 59.6, while all other methods score at or below 50.2. Across all 57 cells, \ours{} achieves the highest overall mean win rate (60.5), outperforming the next-best method, rDPO (55.5), by 5.0 points. Its worst-cell performance is 41.2, closely trailing $\gamma$-PO (42.5). This strong average with a competitive bound demonstrates that our correction recipe generalizes across both clean and noisy regimes rather than overfitting to a specific pattern. Appendix~\ref{app:extended-results} provides per-dataset transfer, cross-dataset noise stress tests, and three-seed runs.

\paragraph{Commercial-model judge validation.}
Open reward models can introduce their own preference biases~\citep{zheng2023llmasajudge,wang2024notfair}. We therefore run a validation with a commercial-model judge on 200 AlpacaEval 2 samples and 80 Vicuna outputs from Qwen2.5-7B. This is intended as an external judging check rather than a replacement for the main evaluation matrix. Table~\ref{tab:commercial-judge-validation} uses Claude Sonnet 4.6~\citep{anthropic2026sonnet46} to report both average single-response quality scores and pairwise win rates against DPO for DPO, rDPO, $\gamma$-PO, ROPO, and \ours{}. Full judging prompts and additional details are provided in Appendix~\ref{app:commercial-judge-validation}.

\subsection{Robustness to Injected Label Noise}
\label{sec:noise-robustness}

We create controlled label noise by swapping chosen and rejected responses with probability $\eta \in \{0.05,0.10,0.20,0.30\}$. Each method is trained on the same corrupted split for each $\eta$.

Table~\ref{tab:noise-robustness} reports the Vicuna evaluation set for Qwen2.5-1.5B using the same fixed \ours{} recipe, with all values measured against the clean one-epoch DPO baseline. \ours{} is strongest at every injected flip rate, including the hardest $\eta=0.30$ setting. ROPO is the closest baseline on this slice, but \ours{} keeps a consistent margin over it from $\eta=0.05$ through $\eta=0.30$. Appendix~\ref{app:noise-radar} reports the corresponding cross-benchmark radar plots for all injected noise rates.

\subsection{Routing Diagnostics}
\label{sec:posterior-diagnostics}

\begin{figure*}[t]
  \centering
  \includegraphics[width=0.80\textwidth]{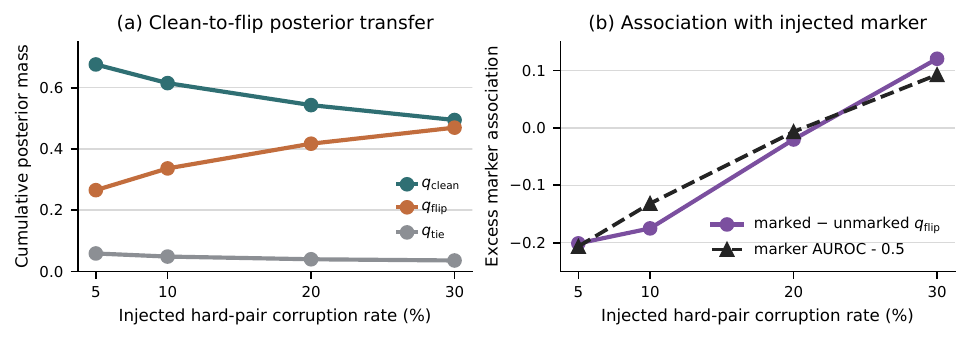}
  \vspace{-10pt}
  \caption{Routing response under hard-pair corruption. Increasing corruption shifts cumulative routing mass from $\pclean$ to $\pflip$, while $\ptie$ remains small because this stress test targets directional reversals rather than low-margin ambiguity. The marker association is a soft routing diagnostic, not label-correction accuracy.}
  \label{fig:posterior-hard-pair}
\end{figure*}

The routing distribution is useful only if its states respond to observable data pathologies. We therefore ask whether controlled hard-pair corruption changes the pair-level routing weights in the expected direction. For each pair, \ours{} recomputes $(\pclean,\pflip,\ptie)$ from the calibrated margin rather than assigning a dataset-level label, so this analysis should be read as a property of the online routing estimator rather than as a separate detector. Importantly, $\pflip$ is not an oracle flip label because the unmarked subset can contain natural annotation errors, ambiguous pairs, and length or style artifacts. Instead, we test a softer claim in which higher injected corruption should make \ours{} trust the observed direction less and allocate more mass to the flip-correction state.

Figure~\ref{fig:posterior-hard-pair} shows a clear dose response on Qwen2.5-7B. As $\eta$ increases from 0.05 to 0.30, cumulative $\pclean$ drops from 0.676 to 0.495, while cumulative $\pflip$ rises from 0.265 to 0.470. The same diagnostic also shows that $\ptie$ stays low under this hard-pair corruption stress test, which is expected because the intervention creates directional reversals rather than weak-gap ambiguity. The marker AUROC increases with heavier corruption, indicating that the routing distribution becomes more aligned with the injected corruption marker without treating the marker as a clean ground-truth label. Full numeric diagnostics are in Appendix~\ref{app:additional-diagnostics}.

\subsection{Tie-State Selectivity}
\label{sec:ambiguity}

We introduce a lightweight selectivity diagnostic for the tie state to see if it targets pairs with weak preference signals. The UltraFeedback dataset~\citep{cui2024ultrafeedback} already includes the original response scores used to initially construct the chosen and rejected pairs. We use these pre-existing scores to calculate the absolute score gap and isolate the bottom and top 20\% of held-out pairs. While not a perfect ambiguity oracle, this allows us to check if $\ptie$ increases on pairs initially deemed weak. Table~\ref{tab:ablation-results} independently verifies the tie loss necessity.

\begin{figure}[t]
  \centering
  \IfFileExists{figures/tie_selectivity_ultrafeedback.pdf}{%
    \includegraphics[width=\columnwidth]{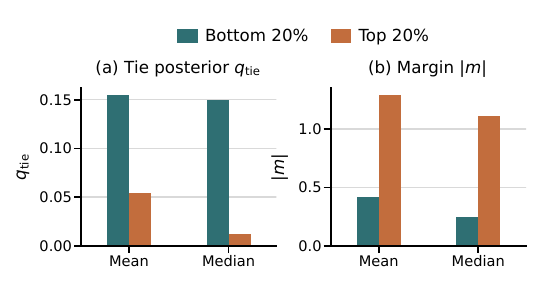}
  }{%
    \fbox{%
      \begin{minipage}[c][0.62\columnwidth][c]{0.92\columnwidth}
        \centering
        \small
        \textbf{UltraFeedback tie-selectivity bar plot unavailable}\\[0.5em]
        Grouped bars with two legend entries, Bottom 20\% and Top 20\%, where slices are defined by the native UltraFeedback chosen--rejected score gap. The x-axis reports mean and median $\ptie$, plus mean and median $|m|$.
      \end{minipage}
    }
  }
  \vspace{-10pt}
  \caption{Tie-state selectivity on held-out UltraFeedback pairs for the clean \ours{}. Bottom 20\% and Top 20\% denote the weakest and strongest-gap pairs based on the native response scores.}
  \label{fig:tie-selectivity}
\end{figure}

Figure~\ref{fig:tie-selectivity} complements the previous corruption analysis. While injected noise shifts routing mass to $\pflip$, we now examine the tie component on naturally weak-gap pairs. We report mean and median values for $\ptie$ and margin magnitude $|m|$ to prevent distortion from outliers. The results demonstrate that $\ptie$ is significantly higher for the Bottom 20\% than the Top 20\%, whereas $|m|$ follows the opposite trend. This confirms the tie state actively suppresses strong updates when the original dataset scores indicate a weak preference.

\begin{table}[t]
  \centering
  \small
  \caption{Injected-noise robustness on Vicuna with Qwen2.5-1.5B, measured by pairwise win rate on the label-flip rate $\eta$ against the clean DPO baseline .}
  \label{tab:noise-robustness}
  \resizebox{\linewidth}{!}{%
  \begin{tabular}{lcccc}
    \toprule
    Method & $\eta=0.05$ & $\eta=0.10$ & $\eta=0.20$ & $\eta=0.30$ \\
    \midrule
    rDPO & 49.38 & 46.88 & 35.00 & 26.88 \\
    $\gamma$-PO & 51.88 & 48.75 & 43.75 & 31.25 \\
    Dr.DPO & 51.25 & 46.25 & 43.12 & 33.75 \\
    ROPO & \underline{55.62} & \underline{55.00} & \underline{65.00} & \underline{56.88} \\
    RE-PO & 46.88 & 48.12 & 40.62 & 38.75 \\
    PLC-DPO & \textbf{66.88} & \textbf{65.62} & \textbf{71.25} & \textbf{61.88} \\
    \bottomrule
  \end{tabular}
  }
  \vspace{-10pt}
\end{table}

\paragraph{Direct tie-state validation.}
We further replace up to 30\% of UltraFeedback preference pairs with exact equal-score pairs and retrain \ours{}, RE-PO, and DPO under these corruption settings. As the injected tie rate increases from 0\% to 30\%, the margin of \ours{} over same-data DPO increases from $+8.3$ to $+18.5$ points. At a 30\% tie rate, \ours{} achieves 68.5, compared with 51.6 for RE-PO. On the independent MultiPref dataset~\citep{zhang2025diverging}, the mean $\ptie$ progressively increases from 0.0811 for unanimous pairs to 0.0904 for divergent pairs and 0.0997 for tie-majority pairs. These results provide evidence that \ours{}'s non-directional mechanism captures synthetic ties and human disagreement. Appendix~\ref{app:tie-validation} provides the full counts, significance tests, and zero-shot per-pair results.

\subsection{Component Ablations}
\label{sec:ablations}

\begin{table*}[t]
  \centering
  \small
  \caption{Component ablations for Qwen2.5-1.5B after one epoch of preference optimization. Values are pairwise win rates against the one-epoch DPO baseline. The full \ours{} row uses the same default recipe reported in Table~\ref{tab:main-results}. Ablation rows use the same recipe and remove or isolate one component while keeping the training surface fixed. We report the evaluation sets where the full method shows the clearest component-level trend.}
  \label{tab:ablation-results}
  \resizebox{\textwidth}{!}{%
  \begin{tabular}{llccccc}
    \toprule
    Variant & Mechanism tested & UFB $\uparrow$ & Alpaca2 $\uparrow$ & Vicuna $\uparrow$ & Evol-Instruct $\uparrow$ & Avg. $\uparrow$ \\
    \midrule
    DPO-LN only & margin rescaling only & 49.48 & 50.56 & 51.25 & 53.35 & 51.16 \\
    Confidence reweight only & downweighting without correction & 50.40 & 52.17 & 58.13 & 52.98 & 53.42 \\
    Remove flip state & no reverse action & 50.12 & 47.33 & 55.62 & 53.42 & 51.62 \\
    Remove tie state & no neutralizing action & 51.95 & \underline{55.65} & \underline{65.62} & 57.02 & \underline{57.56} \\
    Remove EMA calibration & uncalibrated routing scale & \textbf{53.05} & 54.97 & 65.00 & \underline{57.08} & 57.52 \\
    Remove warm-up/gate & no correction deferral & 44.57 & 46.15 & 51.88 & 53.11 & 48.93 \\
    \midrule
    \ours{} full & all correction actions & \underline{52.48} & \textbf{57.70} & \textbf{66.88} & \textbf{58.82} & \textbf{58.97} \\
    \bottomrule
  \end{tabular}
  }
  \vspace{-10pt}
\end{table*}

Table~\ref{tab:ablation-results} isolates the impact of key correction actions and stabilization choices across representative evaluations. DPO-LN tests if gains stem solely from margin rescaling, while confidence reweighting checks if downweighting uncertain examples suffices without reverse or tie actions. Removing the flip or tie state restricts the routing action space, whereas removing EMA calibration or the warm-up gate destabilizes the margin signal. The largest performance drops occur when removing the flip state and warm-up gate, while the tie and EMA variants perform closer to the full method. Hyperparameter details and further diagnostics are provided in Appendices~\ref{app:experimental-details} and~\ref{app:additional-diagnostics}.

\section{Discussion}
\label{sec:discussion}

\paragraph{Why pair-label correction works.}
\ours{} acts on the same pairwise direction that DPO optimizes. This matters because the dominant failure modes of preference data are often directional. A pair can be reliable, reversed, or too ambiguous to support a strong update. The clean/flip/tie decomposition is the smallest latent structure that gives each of these cases a distinct gradient action.

\paragraph{Correction rather than selection.}
Filtering and data-selection methods protect the policy by removing suspicious pairs but often discard valuable signals~\citep{gao2025selectivedpo,liang2025ropo}. A flipped preference pair still contains useful information if the model recognizes the reversal. \ours{} explicitly leverages this distinction by deferring low-confidence examples via a confidence gate, applying a reverse DPO update to likely flipped pairs, and assigning a tie loss to ambiguous cases to discourage arbitrary directional margins.

\paragraph{Why the routing distribution needs stabilization.}
The policy-reference margin is a useful signal, but it is also model-dependent and nonstationary during training. EMA calibration converts the margin into a stream-relative signal. Stop-gradient routing prevents the model from changing the assignment and the objective in the same update. Warm-up and confidence gating keep early low-evidence routing weights from dominating optimization.

\paragraph{Self-confirmation and external agreement.}
Since routing depends on the current policy, early preference errors may reinforce self-confirmation. To mitigate this, we combine a frozen-reference anchor with detached routing weights, EMA calibration, warm-up, and confidence gating. Under 20\% corruption, the final flip-marker AUROC is 0.731, while $\pflip$ predicts disagreement with an independent reward model at 0.779 AUROC on clean pairs. With a 30\% warm start, the final flip-marker AUROC is 0.734. Appendix~\ref{app:self-confirmation} reports the corresponding controls and correction statistics.

\paragraph{The state weights are routing signals.}
The clean, flip, and tie weights should not be interpreted as a ground-truth audit of the dataset. A high $\pflip$ means that, under the current calibrated policy-reference margin and energy scores, the reverse update is the most supported training action for that pair. A high $\ptie$ means that the pair currently provides weak directional evidence for preference optimization. These quantities are useful because they explain how the objective routes gradients, not because they certify clean, flipped, or tied labels independently of the model.

\paragraph{Relation to response-level latent quality.}
A natural alternative is to infer whether each response is good or bad, then route the pair to a response-quality regime. That view is more expressive in some settings, but it also requires an absolute quality signal that DPO does not directly observe. \ours{} makes a narrower choice. It only asks whether the observed pair direction should be reinforced, reversed, or neutralized. This narrower latent variable is easier to connect to DPO's gradient and easier to diagnose with $\pclean$, $\pflip$, and $\ptie$.

\paragraph{Role of the initial state weights.}
Initial state weights are understood as weak anchors for early routing estimates. Once EMA stabilizes and the margin becomes informative, the data-dependent energy scores should dominate these anchors. The recipe sensitivity results in Appendix~\ref{app:plc-recipe-sensitivity} therefore vary the correction strength and routing temperatures together, rather than claiming that any single prior is universally optimal.

\paragraph{When \ours{} reduces to DPO.}
If the dataset is extremely small, the model is severely undertrained, or the routing distribution maintains high entropy, the effective correction weight $\gamma_t C(q)$ remains low. In these scenarios, \ours{} safely reverts to standard DPO. This fallback is preferable to applying confident yet unsupported corrections. Consequently, \ours{} provides the largest gains when sufficient training signals allow the policy-reference margin to become highly informative.

\section{Conclusion}
\label{sec:conclusion}

We introduced \ours{}, a robust DPO objective that treats each observed preference pair label as a latent clean, flip, or tie state. \ours{} estimates an online posterior-like routing distribution from the EMA-calibrated policy-reference margin, stops gradients through the routing weights, and uses them to mix forward DPO, reversed DPO, and tie-regularizing losses. The final objective blends routing correction with standard DPO through warm-up and confidence gating. Across 57 dataset--model--benchmark cells, \ours{} achieves the highest mean win rate and the second-highest worst-cell result among the robust baselines. Injected flip and tie stress tests, independent human-disagreement data, and self-confirmation controls further show that the routing remains stable and assigns distinct actions to reversed and weakly directional pairs. These results support a view of robust preference optimization that corrects the pair label DPO.

\raggedbottom
\section*{Limitations}
\label{sec:limitations}

\ours{} relies on the current policy--reference margin as evidence for the latent label state. Although EMA calibration and our self-confirmation controls mitigate early-stage bias, the margin may still be imperfect when the policy and frozen reference share systematic errors. Performance under distribution shift may therefore benefit from a longer warm-up or more adaptive calibration. Our results empirically demonstrate the effectiveness of the proposed routing mechanism across a range of noisy and ambiguous preference settings. A complementary direction for future work is to characterize further its theoretical properties, including routing error, EMA adaptation, and gating behavior under jointly evolving policy and routing dynamics. Establishing such guarantees would deepen understanding of the mechanism's robustness.

\clearpage
\flushbottom
\section*{Acknowledgments}

This work was supported by Institute for Information \& communications Technology Planning \& Evaluation(IITP)grant funded by the Korea government(MSIT) (RS-2019-II190075, Artificial Intelligence Graduate School Program(KAIST), 10\%), the Institute of Information \& communications Technology Planning \& Evaluation (IITP) grant funded by the Korean government(MSIT) (No. RS-2024-00457882, National AI Research Lab Project, 40\%), the Korea Institute of Energy Technology Evaluation and Planning (KETEP) (No. RS-2026-25533837, 20\%), the Institute of Information \& Communications Technology Planning \& Evaluation (IITP) grant funded by the Korea government (MSIT) (IITP-2026)-RS-2026-25614738, AI Star Fellowship Support Program (20\%), and Institute of Information \& communications Technology Planning \& Evaluation (IITP) under the Artificial Intelligence Innovation Human Resources Development (IITP-2026-RS-2026-25548323) grant funded by the Korea government(MSIT) (10\%).

\bibliography{references}

\clearpage
\appendix

\makeatletter
\renewcommand*{\theHsection}{appendix.\Alph{section}}
\renewcommand*{\theHsubsection}{appendix.\Alph{section}.\arabic{subsection}}
\makeatother

\section*{Appendix Contents}
\label{app:contents}

The appendix presents further results first, followed by additional diagnostics, data construction, experimental and reproducibility details, and responsible research information. Appendix~\ref{app:extended-results} reports the additional cross-model, cross-dataset, tie-state, self-confirmation, sensitivity, and runtime results. Appendix~\ref{app:additional-diagnostics} reports cross-benchmark injected-noise radar plots, commercial-model judge details, full routing diagnostic values, and recipe sensitivity. Appendix~\ref{app:data-construction-details} describes the preference data, injected-noise protocol, and weak-gap slices. Appendix~\ref{app:experimental-details} lists training settings, \ours{} hyperparameter recipes, baseline settings, evaluation protocols, hardware, artifact licenses, and the software environment. Appendix~\ref{app:responsible-research-information} reports AI-assistant use and potential risks.

\section{Further Generalization and Results}
\label{app:extended-results}

All results in this section follow the same method and dataset as in the main text. Unless otherwise specified, we report win rates against DPO trained on the same preference data, and we use no benchmark-specific configuration for each method.

\subsection{Dataset Transfer and Seed Stability}
\label{app:dataset-transfer}

Table~\ref{tab:dataset-transfer-qwen} reports the cross-dataset comparison on Qwen2.5-7B~\citep{qwen2025qwen25}. HH-Golden~\citep{cai2023ulma} and HH-RLHF~\citep{bai2022hhrlhf} provide complementary evaluation settings. HH-Golden contains higher-quality preferred responses for HH prompts, while HH-RLHF exposes every method to naturally noisy, older responses. Nectar-60k~\citep{zhu2024starling} and ORPO-mix-40k~\citep{hong2024orpo} test whether correction degrades performance when the preference data is already comparatively clean.

\begin{table}[H]
  \centering
  \small
  \caption{Mean win rates against DPO trained on the corresponding preference datasets for Qwen2.5-7B.}
  \label{tab:dataset-transfer-qwen}
  \setlength{\tabcolsep}{3.5pt}
  \resizebox{\columnwidth}{!}{%
  \begin{tabular}{lcccc}
    \toprule
    Method & HH-Golden & Nectar-60k & ORPO-mix-40k & HH-RLHF \\
    \midrule
    rDPO & \textbf{78.5} & 46.7 & 48.0 & 52.0 \\
    RE-PO & 54.1 & 48.5 & 48.8 & 46.7 \\
    ROPO & 31.2 & 26.0 & 48.9 & \underline{53.1} \\
    \ours{} & \underline{72.7} & \textbf{50.0} & \textbf{50.9} & \textbf{56.5} \\
    \bottomrule
  \end{tabular}}
\end{table}

On HH-Golden, \ours{} improves over DPO by 22.7 points for Qwen2.5-7B, 23.1 for Llama-3-8B~\citep{grattafiori2024llama3}, and 30.4 for Mistral-7B~\citep{jiang2023mistral7b}. For Mistral-7B, \ours{} reaches 80.4 compared with 78.7 for rDPO. On HH-RLHF, we additionally evaluate intermediate checkpoints throughout training. All methods eventually degrade as training responses fall below the SFT policy's quality, but \ours{} degrades substantially later. At roughly 45\% of training, its in-domain win rate remains at 87.8, compared with 62.0--73.8 for the baselines, and \ours{} is the only method whose in-domain reward remains above the SFT starting point at its best checkpoint before degradation. Table~\ref{tab:seed-stability} reports the corresponding three-seed stability results on clean UltraFeedback~\citep{cui2024ultrafeedback}.

\begin{table}[H]
  \centering
  \small
  \caption{Stability across three random seeds on clean UltraFeedback dataset. Models trained with seeds 42--44 are evaluated against a fixed seed-42 DPO reference, and results are reported as mean $\pm$ standard deviation.}
  \label{tab:seed-stability}
  \setlength{\tabcolsep}{4pt}
  \resizebox{\columnwidth}{!}{%
  \begin{tabular}{lccc}
    \toprule
    Method & AlpacaEval 2 & UltraFeedback & Vicuna \\
    \midrule
    DPO & $47.6\!\pm\!2.7$ & $46.9\!\pm\!3.1$ & $41.3\!\pm\!7.8$ \\
    ROPO & $\underline{58.1\!\pm\!0.9}$ & $\mathbf{55.8\!\pm\!0.5}$ & $\underline{58.5\!\pm\!2.5}$ \\
    \ours{} & $\mathbf{59.4\!\pm\!1.5}$ & $\underline{54.8\!\pm\!1.6}$ & $\mathbf{66.3\!\pm\!7.0}$ \\
    \bottomrule
  \end{tabular}}
\end{table}

\subsection{Full 57-Cell Generalization Table}
\label{app:full-57-cell-results}

Tables~\ref{tab:full-57-ultrafeedback}--\ref{tab:full-57-orpomix} report the 57 cell-level results across UltraFeedback~\citep{cui2024ultrafeedback}, HH-Golden~\citep{cai2023ulma}, Nectar-60k~\citep{zhu2024starling}, and ORPO-mix-40k~\citep{hong2024orpo}. Table~\ref{tab:displayed-57-cell-aggregate} summarizes their per-dataset and overall means.

\begin{table*}[p]
  \centering
  \footnotesize
  \caption{Clean UltraFeedback cell-level results across three base models. Each entry is the final win rate against DPO trained on the same preference data. AE, AE2, Evol, HH, MT, and UF denote AlpacaEval, AlpacaEval 2, Evol-Instruct, HH-RLHF, MT-Bench, and UltraFeedback.}
  \label{tab:full-57-ultrafeedback}
  \setlength{\tabcolsep}{5.2pt}
  \begin{tabular}{llrrrrrr}
    \toprule
    Model & Benchmark & rDPO & RE-PO & Dr.DPO & $\gamma$-PO & ROPO & \ours{} \\
    \midrule
    \multirow{7}{*}{Qwen2.5-7B} & AE & 44.5 & 41.5 & 47.5 & 48.5 & \textbf{58.7} & \underline{58.1} \\
    & AE2 & 48.8 & 42.5 & 47.3 & 51.4 & \underline{58.9} & \textbf{61.4} \\
    & Evol & 46.2 & 44.2 & 47.1 & 50.5 & \underline{59.9} & \textbf{63.0} \\
    & HH & 43.4 & 36.5 & 37.6 & 46.0 & \underline{48.3} & \textbf{65.4} \\
    & MT & 53.1 & 51.2 & 50.6 & 52.5 & \underline{58.8} & \textbf{59.4} \\
    & UF & 47.3 & 43.8 & 48.7 & 50.9 & \underline{58.6} & \textbf{58.8} \\
    & Vicuna & 55.0 & 48.1 & 53.1 & 51.2 & \underline{71.9} & \textbf{76.9} \\
    \midrule
    \multirow{7}{*}{Llama-3-8B} & AE & 43.9 & 49.8 & 53.0 & 50.6 & \textbf{64.2} & \underline{58.5} \\
    & AE2 & 44.1 & 49.0 & 51.6 & 49.4 & \textbf{63.4} & \underline{60.1} \\
    & Evol & 43.1 & 50.6 & 52.2 & 50.6 & \textbf{57.7} & \underline{53.5} \\
    & HH & 48.3 & 56.1 & 54.9 & 54.4 & \underline{56.5} & \textbf{63.3} \\
    & MT & 45.0 & 47.5 & \underline{55.0} & 48.1 & 53.1 & \textbf{63.8} \\
    & UF & 45.0 & 53.0 & 50.8 & 49.2 & \underline{56.6} & \textbf{56.8} \\
    & Vicuna & 36.9 & 52.5 & 60.6 & 45.6 & \textbf{71.2} & \underline{62.5} \\
    \midrule
    \multirow{7}{*}{Mistral-7B} & AE & 45.1 & 47.5 & 48.9 & 48.0 & \textbf{59.4} & \underline{57.9} \\
    & AE2 & 45.2 & 47.3 & 47.1 & 46.5 & \textbf{59.3} & \underline{58.4} \\
    & Evol & 44.3 & 49.3 & 45.7 & 47.4 & \underline{55.4} & \textbf{55.6} \\
    & HH & 48.5 & 52.4 & 50.6 & 52.5 & \underline{57.0} & \textbf{64.2} \\
    & MT & 36.9 & 48.8 & \textbf{56.2} & 48.1 & 54.4 & \underline{55.0} \\
    & UF & 48.9 & 50.5 & 49.7 & 48.5 & \underline{54.5} & \textbf{55.6} \\
    & Vicuna & 40.0 & 46.2 & 46.9 & 52.5 & \textbf{73.8} & \underline{67.5} \\
    \bottomrule
  \end{tabular}
\end{table*}

\begin{table*}[p]
  \centering
  \footnotesize
  \caption{HH-Golden cell-level results across the evaluated model--benchmark pairs.}
  \label{tab:full-57-hhgolden}
  \setlength{\tabcolsep}{5.2pt}
  \begin{tabular}{llrrrrrr}
    \toprule
    Model & Benchmark & rDPO & RE-PO & Dr.DPO & $\gamma$-PO & ROPO & \ours{} \\
    \midrule
    \multirow{7}{*}{Qwen2.5-7B} & AE & \textbf{79.4} & 51.6 & 53.4 & 51.0 & 30.1 & \underline{72.4} \\
    & AE2 & \textbf{78.5} & 51.2 & 51.5 & 52.5 & 31.5 & \underline{73.2} \\
    & Evol & \textbf{72.4} & 51.1 & 52.2 & 52.1 & 34.0 & \underline{67.5} \\
    & HH & \textbf{88.9} & 62.3 & 65.2 & 66.4 & 35.2 & \underline{71.0} \\
    & MT & \underline{70.6} & 50.0 & 50.6 & 57.5 & 31.9 & \textbf{76.9} \\
    & UF & \textbf{76.0} & 50.5 & 50.0 & 49.5 & 32.0 & \underline{71.5} \\
    & Vicuna & \textbf{83.8} & 61.9 & 56.9 & 48.1 & 23.8 & \underline{67.5} \\
    \midrule
    \multirow{4}{*}{Llama-3-8B} & AE2 & \textbf{82.0} & 46.8 & 54.6 & 50.3 & 40.5 & \underline{74.5} \\
    & HH & \textbf{67.6} & 43.4 & 50.4 & 42.5 & 56.1 & \underline{61.2} \\
    & UF & \textbf{81.2} & 48.7 & 49.8 & 52.3 & 40.5 & \underline{71.0} \\
    & Vicuna & \textbf{88.8} & 40.6 & 50.6 & 43.1 & 33.8 & \underline{85.6} \\
    \midrule
    \multirow{4}{*}{Mistral-7B} & AE2 & \textbf{74.0} & 50.9 & 52.0 & 51.8 & 46.0 & \underline{73.3} \\
    & HH & \underline{98.9} & 48.8 & 50.0 & 49.7 & 52.0 & \textbf{99.4} \\
    & UF & \underline{75.5} & 50.1 & 53.4 & 50.0 & 48.3 & \textbf{76.3} \\
    & Vicuna & \underline{66.2} & 48.8 & 50.0 & 51.2 & 49.4 & \textbf{72.5} \\
    \bottomrule
  \end{tabular}
\end{table*}

\begin{table*}[p]
  \centering
  \footnotesize
  \caption{Nectar-60k cell-level results across the evaluated model--benchmark pairs.}
  \label{tab:full-57-nectar}
  \setlength{\tabcolsep}{5.2pt}
  \begin{tabular}{llrrrrrr}
    \toprule
    Model & Benchmark & rDPO & RE-PO & Dr.DPO & $\gamma$-PO & ROPO & \ours{} \\
    \midrule
    \multirow{7}{*}{Qwen2.5-7B} & AE & 43.9 & \textbf{49.6} & 48.8 & 48.6 & 25.1 & \underline{48.9} \\
    & AE2 & 46.1 & 47.8 & 49.2 & \textbf{50.5} & 25.1 & \underline{50.2} \\
    & Evol & 47.0 & 48.9 & \underline{49.9} & \underline{49.9} & 43.9 & \textbf{51.4} \\
    & HH & 45.2 & \textbf{49.7} & 45.2 & 45.3 & 12.7 & \underline{47.3} \\
    & MT & 45.6 & \textbf{53.8} & \underline{51.2} & 50.6 & 22.5 & 41.2 \\
    & UF & 49.4 & 48.9 & \underline{50.8} & 50.1 & 37.5 & \textbf{51.0} \\
    & Vicuna & \textbf{49.4} & 41.2 & 38.8 & 42.5 & 15.0 & \underline{48.8} \\
    \midrule
    \multirow{4}{*}{Llama-3-8B} & AE2 & \underline{50.1} & 48.5 & 49.4 & 48.7 & 49.1 & \textbf{50.6} \\
    & HH & \underline{53.5} & 50.4 & 50.2 & 48.5 & 40.5 & \textbf{59.7} \\
    & UF & 50.1 & \underline{51.0} & 49.8 & 49.9 & \textbf{53.1} & \underline{51.0} \\
    & Vicuna & \textbf{64.4} & \underline{61.2} & 56.2 & 50.6 & 50.6 & 51.2 \\
    \midrule
    \multirow{3}{*}{Mistral-7B} & AE2 & \textbf{52.0} & 48.0 & 49.6 & 48.4 & 45.3 & \underline{50.8} \\
    & HH & 45.9 & \underline{50.4} & 45.7 & 48.9 & 24.8 & \textbf{52.2} \\
    & Vicuna & 45.6 & 48.8 & 44.4 & \underline{50.0} & 38.8 & \textbf{53.8} \\
    \bottomrule
  \end{tabular}
\end{table*}

\begin{table*}[p]
  \centering
  \footnotesize
  \caption{ORPO-mix-40k cell-level results with Qwen2.5-7B.}
  \label{tab:full-57-orpomix}
  \setlength{\tabcolsep}{5.2pt}
  \begin{tabular}{llrrrrrr}
    \toprule
    Model & Benchmark & rDPO & RE-PO & Dr.DPO & $\gamma$-PO & ROPO & \ours{} \\
    \midrule
    \multirow{7}{*}{Qwen2.5-7B} & AE & 45.6 & \textbf{51.2} & \underline{50.2} & 49.0 & 50.1 & 49.5 \\
    & AE2 & 49.8 & 49.8 & \textbf{51.1} & 48.5 & \underline{50.4} & \underline{50.4} \\
    & Evol & 46.7 & 49.4 & 50.1 & 50.0 & \textbf{53.2} & \underline{51.1} \\
    & HH & \textbf{54.1} & 49.4 & \underline{51.9} & 49.9 & 38.8 & 49.4 \\
    & MT & 40.0 & 43.8 & 40.0 & 44.4 & \textbf{49.4} & \underline{48.8} \\
    & UF & 45.8 & 47.8 & \underline{48.8} & 48.6 & \textbf{49.8} & \textbf{49.8} \\
    & Vicuna & \textbf{54.4} & 50.0 & \underline{52.5} & 51.9 & 50.6 & \underline{52.5} \\
    \bottomrule
  \end{tabular}
\end{table*}

\begin{table*}[p]
  \centering
  \small
  \caption{Mean win rates across the 57 cells reported in Tables~\ref{tab:full-57-ultrafeedback}--\ref{tab:full-57-orpomix}. Each dataset column is the mean over its cells, and the overall column gives equal weight to each of the 57 cells.}
  \label{tab:displayed-57-cell-aggregate}
  \setlength{\tabcolsep}{7pt}
  \begin{tabular}{lccccc}
    \toprule
    Method & UltraFeedback & HH-Golden & Nectar-60k & ORPO-mix-40k & Overall \\
           & ($n=21$) & ($n=15$) & ($n=14$) & ($n=7$) & ($n=57$) \\
    \midrule
    rDPO & 45.4 & \textbf{78.9} & 49.2 & 48.1 & \underline{55.5} \\
    RE-PO & 48.0 & 50.4 & \underline{49.9} & 48.8 & 49.2 \\
    Dr.DPO & 50.2 & 52.7 & 48.5 & \underline{49.2} & 50.3 \\
    $\gamma$-PO & 49.6 & 51.2 & 48.7 & 48.9 & 49.7 \\
    ROPO & \underline{59.6} & 39.0 & 34.6 & 48.9 & 46.7 \\
    \ours{} & \textbf{60.7} & \underline{74.3} & \textbf{50.6} & \textbf{50.2} & \textbf{60.5} \\
    \bottomrule
  \end{tabular}
\end{table*}

Across the 57 cells, \ours{} obtains the highest overall win rate at 60.5, followed by rDPO at 55.5. The per-dataset means also show that \ours{} is strongest on UltraFeedback, Nectar-60k, and ORPO-mix-40k, while rDPO is strongest on HH-Golden.

\clearpage

\subsection{Cross-Dataset Label-Noise Stress Tests}
\label{app:cross-dataset-noise}

Table~\ref{tab:cross-dataset-noise} shows that, on clean Nectar-60k and ORPO-mix-40k, \ours{} performs comparably to DPO. Its advantage becomes more pronounced as directional corruption increases, reaching 72.3 and 69.5 at 30\% noise. This clean-to-noisy transition supports the intended adaptive behavior, with correction remaining limited when the observed direction is reliable and strengthening as that direction becomes less trustworthy.

\begin{table}[H]
  \centering
  \scriptsize
  \caption{Three-benchmark mean win rate against same-data DPO on clean and label-flipped Nectar-60k and ORPO-mix-40k with Qwen2.5-7B.}
  \label{tab:cross-dataset-noise}
  \setlength{\tabcolsep}{2.0pt}
  \resizebox{\columnwidth}{!}{%
  \begin{tabular}{lcccccc}
    \toprule
    & \multicolumn{3}{c}{Nectar-60k} & \multicolumn{3}{c}{ORPO-mix-40k} \\
    \cmidrule(lr){2-4}\cmidrule(lr){5-7}
    Method & Clean & 20\% noise & 30\% noise & Clean & 20\% noise & 30\% noise \\
    \midrule
    rDPO & \underline{48.3} & 44.4 & 48.1 & 50.0 & 40.2 & 41.5 \\
    Dr.DPO & 46.2 & \underline{51.8} & 54.3 & \underline{50.8} & 50.6 & 51.9 \\
    RE-PO & 46.0 & 49.4 & 55.1 & 49.2 & 49.4 & 49.8 \\
    ROPO & 25.9 & \textbf{56.5} & \underline{69.3} & 50.3 & \textbf{62.0} & \underline{68.7} \\
    $\gamma$-PO & 47.7 & 48.0 & 50.3 & 49.7 & 48.6 & 51.1 \\
    \ours{} & \textbf{50.0} & \textbf{56.5} & \textbf{72.3} & \textbf{50.9} & \underline{60.5} & \textbf{69.5} \\
    \bottomrule
  \end{tabular}}
\end{table}

\subsection{Direct Tie-State Validation}
\label{app:tie-validation}

\begin{table}[H]
  \centering
  \small
  \caption{Injected-tie stress test. Values are the mean over AlpacaEval 2, UltraFeedback, and Vicuna against DPO trained on the same corrupted data.}
  \label{tab:tie-injection-performance}
  \setlength{\tabcolsep}{4pt}
  \resizebox{\columnwidth}{!}{%
  \begin{tabular}{cccc}
    \toprule
    Tie rate & RE-PO & \ours{} & \ours{} margin over DPO \\
    \midrule
    0\% & \underline{46.8} & \textbf{58.3} & $+8.3$ \\
    5\% & \underline{45.1} & \textbf{60.7} & $+10.7$ \\
    10\% & \underline{44.9} & \textbf{59.7} & $+9.7$ \\
    20\% & \underline{51.3} & \textbf{64.4} & $+14.4$ \\
    30\% & \underline{51.6} & \textbf{68.5} & $+18.5$ \\
    \bottomrule
  \end{tabular}}
\end{table}

Table~\ref{tab:tie-injection-performance} shows that, although the absolute performance of \ours{} is not monotonic across all injection rates, its margin over DPO increases consistently from $+8.3$ to $+18.5$ points. At 5\% tie injection, routing reduces the effective directional signal, measured by $\pclean-\pflip$, on injected pairs by a factor of 4.5. This behavior distinguishes the three-state objective from two-state posterior correction, which must assign each equal-score pair to one of the two preference directions.

\begin{table}[H]
  \centering
  \small
  \caption{Zero-shot per-pair routing on the 30\% exact-tie set using the clean UltraFeedback checkpoint. All Mann--Whitney tests satisfy $p<10^{-15}$.}
  \label{tab:tie-per-pair}
  \setlength{\tabcolsep}{7pt}
  \begin{tabular}{lccc}
    \toprule
    Routing score & Injected & Clean & AUROC \\
    \midrule
    $\pflip$ & 0.194 & 0.127 & 0.659 \\
    $\ptie$ & 0.094 & 0.084 & 0.619 \\
    $\pclean$ & 0.711 & 0.789 & 0.659 \\
    \bottomrule
  \end{tabular}
\end{table}

\begin{table}[H]
  \centering
  \small
  \caption{Zero-shot tie routing on MultiPref human-agreement groups. Both non-unanimous comparisons satisfy $p<10^{-15}$.}
  \label{tab:multipref-routing}
  \setlength{\tabcolsep}{4pt}
  \begin{tabular}{lrcc}
    \toprule
    Group & $n$ & Mean $\ptie$ & vs. unanimous \\
    \midrule
    Unanimous & 4,413 & 0.0811 & --- \\
    Divergent & 4,774 & 0.0904 & $+11.6\%$ \\
    Tie-majority & 1,274 & 0.0997 & $+23.1\%$ \\
    \bottomrule
  \end{tabular}
\end{table}

As shown in Tables~\ref{tab:tie-per-pair} and~\ref{tab:multipref-routing}, the tie channel generalizes beyond the weak-gap proxy considered in the main paper and transfers to human disagreement in MultiPref~\citep{zhang2025diverging}. Its behavior is graded rather than binary, with $\ptie$ highest when human annotators favor a tie, intermediate under split preferences, and lowest under unanimous preferences.

\subsection{Self-Confirmation Controls}
\label{app:self-confirmation}

Under 20\% label-flip corruption, the final cumulative flip-marker AUROC reaches 0.731. The mean $\pflip$ is 0.473 on injected flips and 0.197 on unmodified pairs, indicating that the routing signal assigns substantially higher flip probability to corrupted examples. On 8,000 held-out clean UltraFeedback pairs, $\pflip$ also predicts disagreement between the dataset label and an independent reward model~\citep{liu2026skyworkrewardv2} with an AUROC of 0.779 and a reported range of $[0.766, 0.791]$. The reward model disagrees with the dataset label on 22.9\% of pairs, and the strength of this disagreement correlates with $\pflip$ at a Spearman coefficient of 0.561.

We further extend the warm-up for \ours{} and DPO from 7\% to 30\% on the 20\%-corrupted data. The resulting cumulative flip-marker AUROC remains comparable at 0.734 versus 0.731, while the three-benchmark mean improves from 71.0 to 73.6. Correction is also limited during early training. The first 20\% of training accounts for only 1.9\% of the total correction weight, with a mean correction weight of 0.026 during this period compared with 0.286 over the full run. The noisy setting exhibits the same 1.9\% early-training share.

\subsection{One-Dimensional Sensitivity Sweeps}
\label{app:full-hyperparameter-sweep}

\begin{table*}[t]
  \centering
  \small
  \caption{One-dimensional retraining sweeps on Qwen2.5-7B. Performance is the three-benchmark mean over UltraFeedback, AlpacaEval 2, and Vicuna against DPO.}
  \label{tab:full-hyperparameter-sweep}
  \setlength{\tabcolsep}{5pt}
  \resizebox{0.70\textwidth}{!}{%
  \begin{tabular}{llll}
    \toprule
    Hyperparameter & Grid & Win rate at each grid point & Range \\
    \midrule
    $\tau_{\mathrm{dir}}$ & 0.35 / 0.75 / 1.0 / 1.5 & 60.1 / 55.7 / 52.9 / 56.5 & $[52.9,60.1]$ \\
    $\tau_{\mathrm{tie}}$ & 0.85 / 1.0 / 1.3 / 1.5 & 60.6 / 59.1 / 60.9 / 62.9 & $[59.1,62.9]$ \\
    $\kappa$ & 0.5 / 1.0 / 1.5 / 2.0 & 60.3 / 58.2 / 60.3 / 60.8 & $[58.2,60.8]$ \\
    $\gamma_{\max}$ & 0.25 / 0.5 / 0.7 / 0.95 & 55.0 / 55.5 / 55.3 / 62.4 & $[55.0,62.4]$ \\
    $\rho_{\mathrm{warm}}$ & 0.0 / 0.03 / 0.15 / 0.30 & 57.9 / 61.6 / 64.4 / 59.1 & $[57.9,64.4]$ \\
    $\alpha$ (EMA) & 0.90 / 0.95 / 0.99 / 0.995 & 58.7 / 61.8 / 61.9 / 60.2 & $[58.7,61.9]$ \\
    \bottomrule
  \end{tabular}}
\end{table*}

Table~\ref{tab:full-hyperparameter-sweep} shows that all 24 configurations outperform the DPO reference of 50.0, with an overall mean of 59.3 and a range of 52.9--64.4. The fixed default configuration achieves 58.3, indicating that a single favorable setting does not drive the improvement. Among the six hyperparameters, $\tau_{\mathrm{dir}}$ and $\gamma_{\max}$ produce the largest variation in performance.

\subsection{Computational Overhead}
\label{app:computational-overhead}

\begin{table}[H]
  \centering
  \small
  \caption{Runtime comparison under identical training configurations on eight NVIDIA B200 GPUs. Total wall-clock time is measured over a 110-step run, while median seconds per step are computed over three repeated 500-step windows. Brackets denote the range across the three windows.}
  \label{tab:computational-overhead}
  \setlength{\tabcolsep}{4pt}
  \resizebox{\columnwidth}{!}{%
  \begin{tabular}{lccc}
    \toprule
    Method & 110-step total & Median s/step & Overhead \\
    \midrule
    DPO & 378 s & 1.362 $[1.312, 1.430]$ & --- \\
    \ours{} & 377 s & 1.110 $[1.068, 1.190]$ & $\approx 0$ \\
    \bottomrule
  \end{tabular}}
\end{table}

Table~\ref{tab:computational-overhead} shows that \ours{} reuses the policy and reference log probabilities already computed for the DPO loss, requiring no additional forward or backward passes and no additional model copies. It therefore has the same model-memory requirements as DPO, adding only $O(\text{batch})$ scalar routing operations and an EMA update at each step. Under identical configurations on eight NVIDIA B200 GPUs, the measured wall-clock time is comparable to DPO. Although the per-step medians are lower for \ours{}, this difference may reflect variability in the shared dataloader and logging pipeline rather than a systematic speedup. We therefore conclude only that \ours{} is not slower than DPO in our measurements.

\section{Additional Diagnostics and Sensitivity Checks}
\label{app:additional-diagnostics}

\subsection{Cross-Benchmark Injected-Noise Radar}
\label{app:noise-radar}

\begin{figure*}[p]
  \centering
  \includegraphics[width=\textwidth]{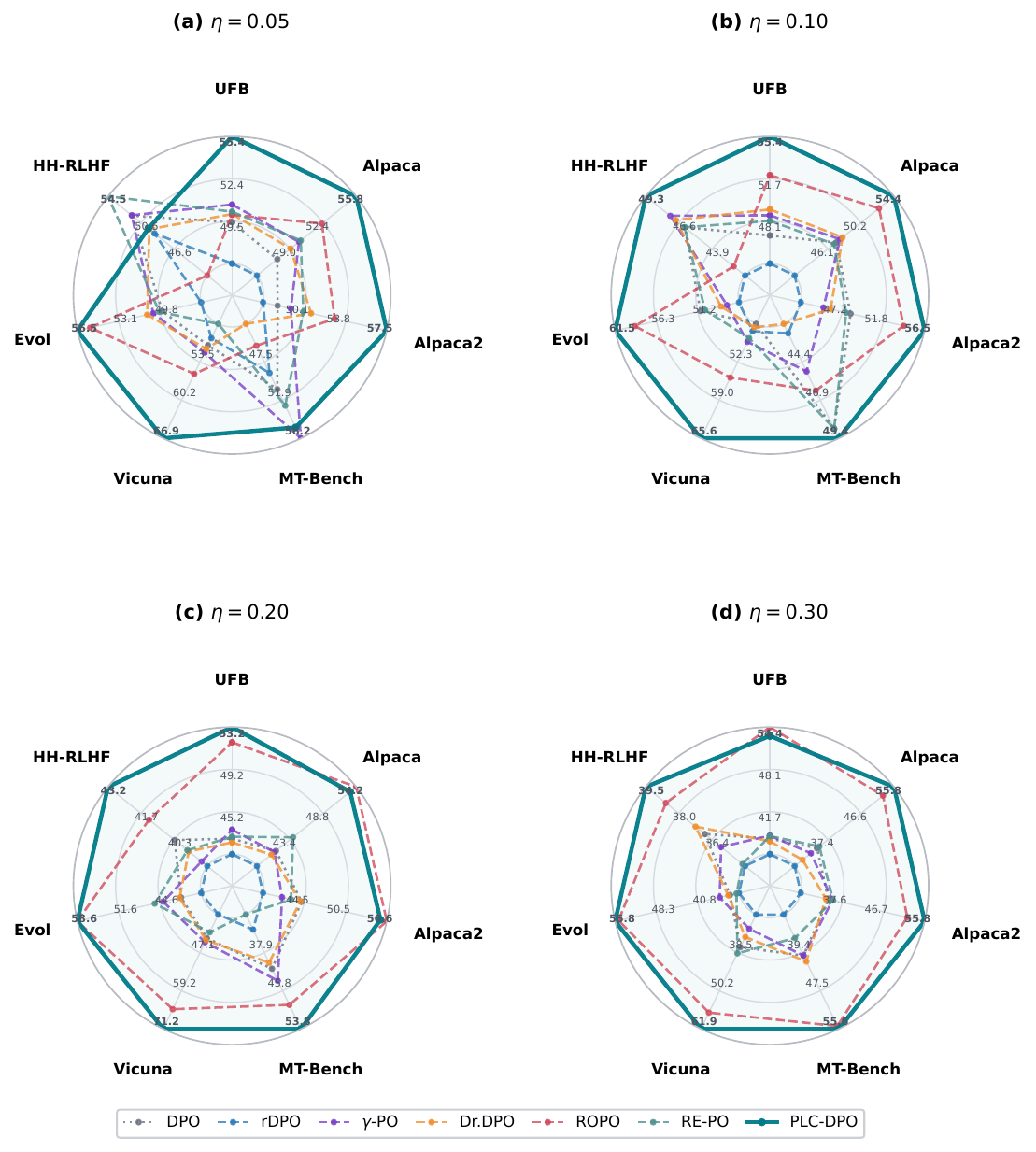}
  \caption{Cross-benchmark injected-noise robustness for Qwen2.5-1.5B after one epoch of preference optimization, measured as pairwise win rate against the clean one-epoch DPO baseline. Each radar axis corresponds to an evaluation set, and each panel corresponds to a label-flip rate. Axis-specific rings are linearly scaled from the smallest to the largest plotted method value for that dataset and noise rate, with the outer-ring value shown in bold. \ours{} uses the fixed default recipe for all axes.}
  \label{fig:noise-robustness-radar}
\end{figure*}

Figure~\ref{fig:noise-robustness-radar} expands Table~\ref{tab:noise-robustness} from the selected Vicuna view to all evaluation sets for Qwen2.5-1.5B. For \ours{}, each point uses the fixed default recipe rather than selecting the best recipe per noise rate or evaluation set.

\clearpage
\clearpage
\twocolumn[
\begin{minipage}{\textwidth}
  \centering
  \small
  \captionof{table}{Routing response under hard-pair label corruption for Qwen2.5-7B with the aggressive \ours{} recipe. The marker AUROC ranks injected hard-pair flips using $\pflip$. It is a soft routing diagnostic, not oracle label-correction accuracy.}
  \label{tab:posterior-hard-pair}
  \begin{tabular}{cccccccc}
    \toprule
    $\eta$ & $\pclean$ & $\pflip$ & $\ptie$ & Argmax flip & Marked $\pflip$ & Unmarked $\pflip$ & Marker AUROC \\
    \midrule
    0.05 & 0.676 & 0.265 & 0.059 & 0.264 & 0.074 & 0.275 & 0.293 \\
    0.10 & 0.615 & 0.336 & 0.049 & 0.353 & 0.179 & 0.354 & 0.368 \\
    0.20 & 0.543 & 0.417 & 0.040 & 0.442 & 0.401 & 0.421 & 0.493 \\
    0.30 & 0.495 & 0.470 & 0.036 & 0.498 & 0.554 & 0.434 & 0.592 \\
    \bottomrule
  \end{tabular}
\end{minipage}

\vspace{\dbltextfloatsep}
]

\subsection{Full Routing Diagnostic Values}

Table~\ref{tab:posterior-hard-pair} gives the numeric values behind Figure~\ref{fig:posterior-hard-pair}. The table is included here because the main figure emphasizes the qualitative routing trend, while the numeric view makes the movement of $\pclean$, $\pflip$, and $\ptie$ across noise rates explicit. The marker AUROC ranks injected hard-pair flips using $\pflip$ and is interpreted only as a soft routing diagnostic. It is not an estimate of oracle label-correction accuracy because unmarked pairs can still contain natural annotation errors, ambiguous directions, or stylistic artifacts.

\subsection{Commercial-Model Judge Validation Details}
\label{app:commercial-judge-validation}

For the compact main-table validation in Table~\ref{tab:commercial-judge-validation}, we use a commercial-model judge on AlpacaEval 2 and Vicuna outputs from Qwen2.5-7B. The goal is to test whether the main trend survives a different judge source, not to duplicate every open reward-model comparison. We use the same generated responses as in Table~\ref{tab:main-results}, score each response independently with Claude Sonnet 4.6~\citep{anthropic2026sonnet46}, and additionally run pairwise comparisons against the corresponding one-epoch DPO baseline. For pairwise judging, we randomize response order before judging and map ties to half wins. The full judge prompt, model version, decoding settings, and per-dataset sample ids are kept fixed across methods.

For single-answer quality diagnostics, the script stores one augmented JSONL record per response and a companion id-to-score JSON file, so these scalar scores can be averaged, filtered by method, or inspected at the prompt level without converting them into pairwise wins. Figure~\ref{fig:single-answer-judge-prompt} records the exact single-answer prompt used for this scalar-score audit.

\begin{figure*}[t]
  \centering
  \setlength{\fboxsep}{6pt}
  \fbox{%
    \begin{minipage}{0.96\textwidth}
    \small
    \textbf{System.} Please act as an impartial judge and evaluate the quality of the response provided by an AI assistant to the user question displayed below. Your evaluation should consider factors such as the helpfulness, relevance, accuracy, depth, creativity, and level of detail of the response. Begin your evaluation by providing a short explanation. Be as objective as possible. After providing your explanation, please rate the response on a scale of 1 to 10 by emitting the word Rating followed by a bracketed score such as [[5]].

    \vspace{0.5em}
    \textbf{User.} [Question]\\
    \texttt{\{question\}}\\
    {}[The Start of Assistant's Answer]\\
    \texttt{\{answer\}}\\
    {}[The End of Assistant's Answer]
    \end{minipage}%
  }
  \caption{Single-answer LLM-as-a-judge~\citep{zheng2023llmasajudge} grading prompt used for the AlpacaEval2~\cite{li2023alpacaeval} response-quality scores. The judge is asked to provide a short rationale and then emit a bracketed 1--10 rating, which is parsed into per-id scalar scores.}
  \label{fig:single-answer-judge-prompt}
\end{figure*}

\subsection{\ours{} Recipe Sensitivity}
\label{app:plc-recipe-sensitivity}

\begin{figure*}[t]
  \centering
  \includegraphics[width=\textwidth]{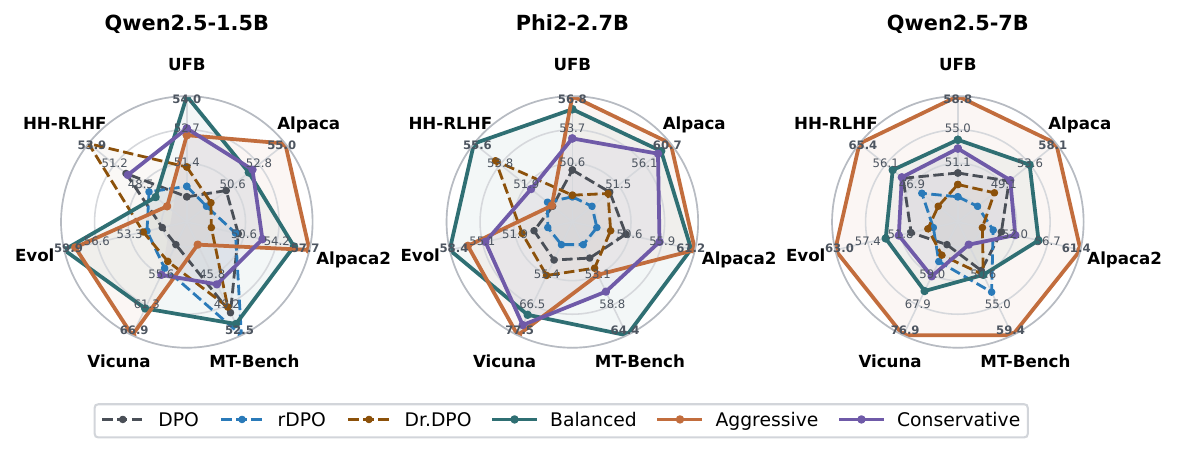}
  \caption{\ours{} recipe sensitivity and baseline comparison after one epoch of preference optimization. Each radar axis is one evaluation set, each panel reports one base model, and each axis is scaled so that the largest selected method value reaches the outer ring.}
  \label{fig:plc-recipe-radar}
\end{figure*}

Figure~\ref{fig:plc-recipe-radar} compares the one-epoch \ours{} recipe variants with DPO, rDPO, and Dr.DPO across evaluation sets. Balanced denotes the unqualified \texttt{plc-dpo} row in the result CSVs, while aggressive and conservative denote the corresponding recipe-specific rows. DPO is fixed at 50 because all win rates are measured against the corresponding one-epoch DPO baseline. This figure explains why the main text uses a fixed aggressive recipe instead of selecting a recipe separately for each model and evaluation set.

\clearpage
\twocolumn

\section{Data Construction Details}
\label{app:data-construction-details}

\subsection{Source Dataset}

The main experiments use the train set from \texttt{HuggingFaceH4/ultrafeedback\_binarized}~\citep{cui2024ultrafeedback}. Each example contains a prompt, an observed chosen response, and an observed rejected response derived from UltraFeedback ratings. We use the provided labels and do not perform offline relabeling.

For the generalization experiments, we additionally train on HH-Golden~\citep{cai2023ulma}, Nectar-60k~\citep{zhu2024starling}, ORPO-mix-40k~\citep{hong2024orpo}, and the original HH-RLHF~\citep{bai2022hhrlhf}. Nectar-60k is a fixed 60,000-pair sample of Nectar, ORPO-mix-40k is the fixed community preference mixture used in our experiments, and HH-Golden replaces the preferred responses of the corresponding HH-RLHF prompts with higher-quality generations. Every objective receives the same processed split within a dataset-model comparison.

\subsection{Synthetic Label-Noise Injection}

For a noise rate $\eta$, we sample each training pair independently with probability $\eta$ and swap the chosen and rejected responses. The rates are $\eta \in \{0,0.05,0.10,0.20,0.30\}$. The random seed is shared across methods so that every method sees the same corrupted dataset at a given noise rate.

\subsection{Weak-Gap Preference Dataset}

For the tie-state selectivity diagnostic, we use held-out preference pairs and rank them by the absolute score gap between the dataset-provided chosen and rejected responses. The low-gap and high-gap slices are the bottom and top 20\% of this ranking, respectively. These slices are used solely to analyze routing-allocation and policy-reference margins for existing preference pairs.

\subsection{Exact-Tie Injection Dataset}

For exact-tie stress tests, we replace 5--30\% of UltraFeedback training pairs with equal-score response pairs in a randomly chosen/rejected orientation. The corruption sets are nested across rates, and we train DPO, RE-PO, and \ours{} from scratch on the same split at each rate. The 30\% set contains 12,012 analysis pairs: 3,567 injected ties and 8,445 unmodified pairs.

\subsection{MultiPref Human-Disagreement Dataset}

For the human-disagreement evaluation, we use the 10,461 preference pairs in MultiPref~\citep{zhang2025diverging} as an independent evaluation set, score them without further training, and preserve their multi-annotator labels. We partition the evaluation pairs into unanimous, divergent, and tie-majority groups according to whether annotators fully agree, express a minority disagreement, or most frequently select an explicit tie.

\section{Experimental and Reproducibility Details}
\label{app:experimental-details}

\subsection{Training Details}

All methods share the same training surface unless a baseline requires an objective-specific hyperparameter. We use a single epoch of preference optimization for each model and keep the optimizer, scheduler, adapter rank, and effective batch size fixed across objectives. This controlled setup isolates the preference objective's effect from changes in training budget or adaptation capacity. We report values that differ across model sizes, such as per-device batch sizes or gradient accumulation steps, separately when applicable.

\begin{table}[t]
\centering
\small
\caption{Common training hyperparameters used across preference-optimization methods unless otherwise stated.}
\label{tab:training-details}
\setlength{\tabcolsep}{4pt}
\begin{tabular}{p{0.32\linewidth}p{0.54\linewidth}}
\toprule
Category & Setting \\
\midrule
Base checkpoints & \texttt{Qwen2.5-1.5B} SFT~\citep{airhl2025qwen2.5-1.5b-ultrachat200k}, \texttt{Phi-2-2.7B} SFT~\citep{lole252024phi2}, and \texttt{Qwen2.5-7B}, \texttt{Llama-3-8B}, and \texttt{Mistral-7B} SFT models~\citep{qwen2025qwen25,grattafiori2024llama3,jiang2023mistral7b,ding2023ultrachat} \\
Main preference data & \path{HuggingFaceH4/ultrafeedback_binarized} train set~\citep{cui2024ultrafeedback,huggingfaceh42023ultrafeedbackbinarized} \\
Transfer preference data & HH-Golden, Nectar-60k, ORPO-mix-40k, and HH-RLHF \\
Training epochs & $1$ \\
Optimizer & AdamW~\citep{loshchilov2018adamw} \\
Learning rate & $1\times 10^{-5}$ \\
Scheduler & Cosine decay with 10\% warm-up \\
Preference scale & $\beta=0.01$ \\
LoRA rank~\citep{hu2022lora} & $16$ \\
LoRA alpha & $16$ \\
Effective batch size & $64$ \\
\bottomrule
\end{tabular}
\end{table}

For the 1.5B and 2.7B models, we obtain the same effective batch size by adjusting only the per-device batch size and gradient accumulation to fit device memory. For the 7B model, we keep the objective-level settings unchanged and modify only memory-dependent implementation choices, such as gradient accumulation and attention backend. All reported methods use the same decoding and evaluation settings after training.

\subsection{\ours{} Recipes}

The main text treats $\tau_{\mathrm{dir}}$, $\tau_{\mathrm{tie}}$, $\gamma_{\max}$, and $\kappa$ as objective-level controls, while $\alpha$ and $\rho_{\mathrm{warm}}$ are calibration and scheduling settings. Here, $\alpha$ is the EMA decay used to update the streaming mean and variance of the detached DPO margin. Larger values make calibration change more slowly, while smaller values make it follow the current training stream more quickly. To avoid benchmark-specific tuning in the main comparisons, we define three preset recipes and use the main-experiment default recipe unless otherwise stated. The initial state prior and numerical standard-deviation floor are fixed implementation constants, so we omit them from the recipe table.

\begin{table}[t]
\centering
\small
\caption{\ours{} hyperparameter recipes. The aggressive recipe is used for the main results unless otherwise stated.}
\label{tab:plc-recipes}
\setlength{\tabcolsep}{4pt}
\begin{tabular}{lccc}
\toprule
Hyperparameter & Conservative & Balanced & Aggressive \\
\midrule
$\alpha$ & 0.995 & 0.99 & 0.98 \\
$\tau_{\mathrm{dir}}$ & 1.00 & 0.75 & 0.55 \\
$\tau_{\mathrm{tie}}$ & 0.85 & 1.00 & 1.15 \\
$\rho_{\mathrm{warm}}$ & 0.15 & 0.10 & 0.07 \\
$\gamma_{\max}$ & 0.50 & 0.70 & 0.85 \\
$\kappa$ & 1.5 & 1.0 & 0.8 \\
\bottomrule
\end{tabular}
\end{table}

\subsection{Baseline Settings}

All baseline runs use the same SFT checkpoint, preference data, optimizer, scheduler, adapter rank, epoch budget, and effective batch size unless the objective itself requires a different reward scale. We follow the official papers, public documentation, and available code for each baseline's objective-specific settings~\citep{wu2025drdpo,liang2025ropo,sun2025gammapo,cao2026repo,ethayarajh2024kto}, while matching the shared hyperparameter setup used for \ours{}.

\paragraph{Noisy-preference baselines.}
We run Dr.DPO, ROPO, $\gamma$-PO, and RE-PO according to their official method descriptions under the same shared training setup described above. KTO-Pair is the only baseline that changes the supervision format. Each preference pair $(y_w,y_l)$ is converted into pointwise feedback by treating $y_w$ as desirable and $y_l$ as undesirable.

This setup keeps data, model, optimizer, adapter capacity, and training budget matched across methods while preserving each baseline's method-specific objective.

\subsection{Evaluation Protocol}

For all benchmarks, we keep decoding parameters and judge configurations strictly identical across all evaluated methods to ensure a fair comparison.

\paragraph{AlpacaEval \& AlpacaEval 2.}
We evaluate our models on both AlpacaEval and AlpacaEval 2~\citep{li2023alpacaeval}, each consisting of 805 test samples. To manage inference costs when using commercial LLM-as-a-judge models (e.g., GPT-4), we use a fixed subset of the first 200 samples, selected by their chronological IDs. We follow the official evaluation protocol and report both the length-controlled win rate and the raw win rate where available.

\paragraph{UltraFeedback.}
We evaluate on the full test split of the UltraFeedback dataset~\citep{cui2024ultrafeedback}, comprising all 2,000 samples.

\paragraph{Vicuna.}
We evaluate on the standard Vicuna benchmark~\citep{chiang2023vicuna}, which consists of 80 test prompts.

\paragraph{HH-RLHF.}
For the HH-RLHF benchmark~\citep{bai2022hhrlhf}, we construct a test subset of 805 instances by extracting the samples with the earliest IDs from the original test split.

\paragraph{MT-Bench.}
We use the standard 80-prompt, two-turn MT-Bench evaluation protocol~\citep{zheng2023llmasajudge} and maintain the same judge configuration across all methods.

\subsection{Hardware and Runtime}
Unless otherwise noted, the main training experiments were conducted using two NVIDIA RTX A6000 GPUs. The computational-overhead experiment in Table~\ref{tab:computational-overhead} was conducted using eight NVIDIA B200 GPUs. Within every direct comparison, all compared methods use the same device type, hardware allocation, parallelism strategy, and evaluation environment. Under the two-A6000 configuration, fine-tuning a 7B parameter model with LoRA on the UltraFeedback dataset (comprising approximately 64k examples) required about 8 hours of total runtime.

\subsection{Models \& Datasets Licenses}
To comply with guidelines for using scientific models and datasets, we detail the licenses of the pre-trained models used in our study. All models are publicly accessible and permit use for academic research. The licenses for the specific models used are as follows:

\begin{itemize}
    \item \texttt{Skywork-Reward-V2-Llama-3.1-8B} is distributed under the Llama 3.1 Community License~\citep{skywork2025rewardv2llama31card}.
    \item \texttt{AIR-hl/Qwen2.5-1.5B-ultrachat200k} is distributed under the Apache License 2.0~\citep{airhl2025qwen2.5-1.5b-ultrachat200k}.
    \item \texttt{lole25/phi-2-sft-ultrachat-full} is distributed under the MIT License~\citep{lole252024phi2}.
    \item \path{HuggingFaceH4/ultrafeedback_binarized} is distributed under the MIT License~\citep{huggingfaceh42023ultrafeedbackbinarized}.
\end{itemize}

The remaining software packages used in our experiments are governed by their respective repository licenses; we report exact package versions below for reproducibility.

\subsection{Software Environment}
To ensure reproducibility, we detail the software environment and library versions used in our study. We conducted all experiments in a virtual environment running Python 3.11.15 and an independent installation of CUDA Toolkit 12.8. The core deep learning framework was PyTorch (v2.10.0+cu128)~\citep{paszke2019pytorch}.

For the loading, fine-tuning, and evaluation of large language models, we extensively relied on the Hugging Face ecosystem, specifically utilizing \texttt{transformers} (v5.2.0)~\citep{wolf2020transformers}, \texttt{peft} (v0.18.1), \texttt{trl} (v0.24.0), \texttt{accelerate} (v1.13.0), and \texttt{datasets} (v4.3.0).

To optimize computational efficiency and memory footprints during training, we integrated \texttt{bitsandbytes} (v0.49.2) for quantization~\citep{dettmers20228bit}, alongside \texttt{deepspeed} (v0.18.8), \texttt{xformers} (v0.0.35), and FlashAttention-2 (\texttt{flash-attn} v2.8.3)~\citep{dao2024flashattention2}. To accelerate LoRA fine-tuning~\citep{hu2022lora}, we used the \texttt{unsloth} and \texttt{unsloth\_zoo} libraries, compiled directly from their latest source repositories. Finally, experiment tracking and model management were handled using Weights \& Biases (\texttt{wandb}) and \texttt{huggingface-hub} (v1.7.1).

\section{Responsible Research Information}
\label{app:responsible-research-information}

\subsection{Use of AI Assistants}

In accordance with the ACL Rolling Review and EMNLP 2026 policies on AI assistant use, we transparently disclose our use of large language models (LLMs) throughout this project.

Specifically, we used AI assistants (such as ChatGPT) during the research and development phase to help write, debug, and optimize the codebase for our computational experiments. In the manuscript preparation stage, AI tools were used to facilitate data visualization by generating scripts for matplotlib to construct paper tables and figures, structure specific content elements, and perform comprehensive grammar checks, proofreading, and editorial reviews to enhance the overall clarity and readability of the text.

We explicitly note that all core scientific ideas, experimental designs, interpretations of the results, and final text revisions remain entirely the original work of the authors, who maintain full accountability for the contents of this publication.

\subsection{Potential Risks}

As we advance the fine-tuning of large language models (LLMs), we acknowledge several inherent ethical and societal risks. Like most current autoregressive LLMs, our models are prone to hallucinations. They may generate fluent but factually incorrect outputs, making them unsuitable for high-stakes domains without rigorous human oversight. Furthermore, because they are trained on public datasets such as UltraFeedback, they may inherit and reproduce implicit societal biases, stereotypes, or toxic language that are naturally present in human-generated data. Finally, the open accessibility of these models and tuning methods carries a dual-use risk of malicious exploitation, such as the automated generation of misinformation or spam. We strongly urge robust safety guardrails and continuous human evaluation before any real-world deployment.

\end{document}